\documentclass{article}

\usepackage{microtype}
\usepackage{graphicx}
\usepackage{subcaption}
\usepackage{booktabs} 

\usepackage{hyperref}

\usepackage{algorithm}
\usepackage{algorithmic}

\usepackage[preprint]{icml2026}

\usepackage{amsmath}
\usepackage{amssymb}
\usepackage{mathtools}
\usepackage{amsthm}

\usepackage{pifont}
\usepackage{multirow}
\usepackage{multicol}
\usepackage{xcolor}
\usepackage[table]{xcolor}
\usepackage{rotfloat}
\usepackage{diagbox}
\hypersetup{
    colorlinks=true,
    citecolor=teal,
    linkcolor=red,
    filecolor=magenta,
    urlcolor=magenta}

\usepackage[capitalize,noabbrev]{cleveref}

\theoremstyle{plain}
\newtheorem{theorem}{Theorem}[section]

\theoremstyle{definition}
\newtheorem{definition}[theorem]{Definition}

\theoremstyle{remark}
\newtheorem{remark}[theorem]{Remark}

\usepackage[textsize=tiny]{todonotes}

\icmltitlerunning{To See Far, Look Close: Evolutionary Forecasting for Long-term Time Series}

\begin{document}

\twocolumn[
  \icmltitle{To See Far, Look Close: Evolutionary Forecasting for Long-term Time Series}
  


  \icmlsetsymbol{equal}{*}

  \begin{icmlauthorlist}
    \icmlauthor{Jiaming Ma}{ustc}
    \icmlauthor{Siyuan Mu}{sicau}
    \icmlauthor{Ruilin Tang}{hnu}
    \icmlauthor{Haofeng Ma}{unnc}
    \icmlauthor{Qihe Huang}{ustc}
    \icmlauthor{Zhengyang Zhou}{ustc}
    \icmlauthor{Pengkun Wang}{ustc}
    \icmlauthor{Binwu Wang}{ustc}
    \icmlauthor{Yang Wang}{ustc}
  \end{icmlauthorlist}

  \icmlaffiliation{ustc}{University of Science and Technology of China}
  \icmlaffiliation{hnu}{Hunan University}
  \icmlaffiliation{unnc}{The University of Nottingham Ningbo China}
  \icmlaffiliation{sicau}{Sichuan Agricultural University}



  \icmlkeywords{Machine Learning, ICML}

  \vskip 0.3in
]



\printAffiliationsAndNotice{}  

\begin{abstract}
  The prevailing Direct Forecasting (DF) paradigm dominates Long-term Time Series Forecasting (LTSF) by forcing models to predict the entire future horizon in a single forward pass. While efficient, this rigid coupling of output and evaluation horizons necessitates computationally prohibitive re-training for every target horizon. In this work, we uncover a counter-intuitive optimization anomaly: models trained on short horizons—when coupled with our proposed Evolutionary Forecasting (EF) paradigm—significantly outperform those trained directly on long horizons. We attribute this success to the mitigation of a fundamental optimization pathology inherent in DF, where conflicting gradients from distant futures cripple the learning of local dynamics. We establish EF as a unified generative framework, proving that DF is merely a degenerate special case of EF. Extensive experiments demonstrate that a singular EF model surpasses task-specific DF ensembles across standard benchmarks and exhibits robust asymptotic stability in extreme extrapolation. This work propels a paradigm shift in LTSF: moving from passive ``Static Mapping'' to autonomous ``Evolutionary Reasoning''.
\end{abstract}


\section{Introduction}\label{sec:intro}

\begin{figure}[!t]
    \centering
    \includegraphics[width=\linewidth]{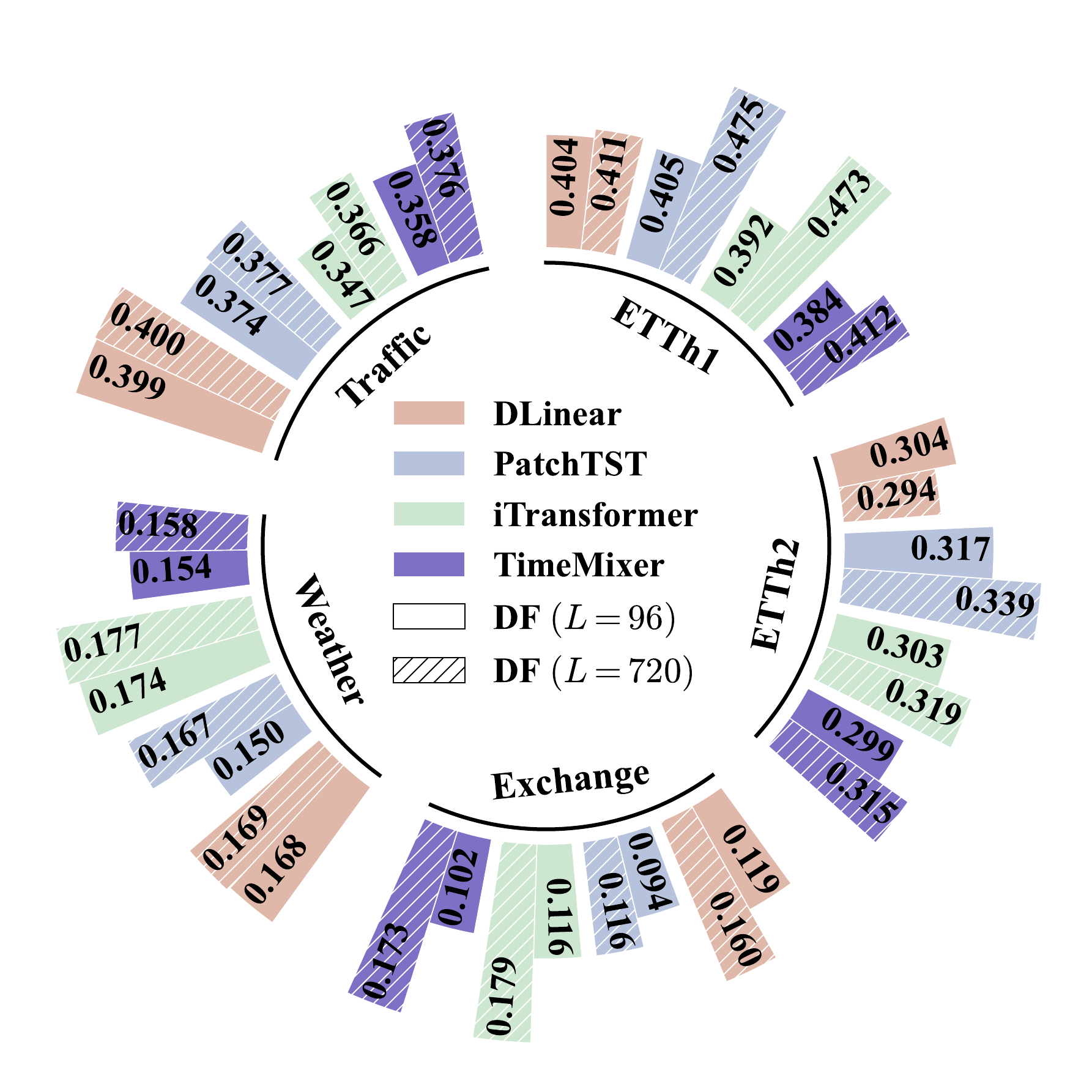}
    \caption{MSE comparison of $96$-horizon forecasting on common real-world datasets. Using a fixed input length $T=720$, we compare models trained directly with an output horizon $L$ matching the evaluation horizon ($L=96$, solid bars) against models trained with a longer output horizon ($L=720$, hatched bars). For the latter, predictions are obtained by truncating the long series to match $96$. Lower bars indicate better performance. The results demonstrate that truncation leads to sub-optimal performance compared to directly training for the target horizon.}
    \label{fig:first_truncated}
\end{figure}

Long-term Time Series Forecasting (LTSF) is a fundamental task across a diverse spectrum of critical domains, encompassing power grid management, financial market analysis, intelligent transportation systems, extreme weather forecasting and proactive healthcare monitoring \cite{huang2023crossgnn,qiu2024tfb,huang2024hdmixer,ma2025causal,ma2025bist}. The deep learning methodologies in this field are currently dominated by the Direct Forecasting (DF) paradigm~\cite{zhou2021informer}, where models' input length and output horizon serve as parameters, outputting all future multi-step predictions in a single forward process, effectively sidestepping the catastrophic error accumulation inherent in traditional step-wise recursion \cite{box2015time,salinas2020deepar,lai2018modeling,li2019enhancing,child2019generating}. Standard evaluation protocols in LTSF typically assess models across multiple horizons to gauge performance at varying temporal scales~\cite{wu2022timesnet,liang2022basicts}. Intuitively, a long-horizon prediction inherently encapsulates shorter segments (e.g., a 720-step forecast naturally includes the 96-step prefix). However, a prevailing empirical consensus within the community reveals a critical limitation: simply truncating a long output series yields drastically inferior performance on the initial segment compared to a model exclusively trained for that specific short horizon. This performance degradation has been systematically visualized in Figure~\ref{fig:first_truncated}. Consequently, researchers implicitly default to strictly coupling the model's output horizon with the task's evaluation horizon~\cite{qiu2024tfb}. This rigid practice enforces a computationally prohibitive benchmarking protocol, necessitating the re-customization and re-training of separate models for every distinct evaluation horizon.

Crucially, this expensive consensus overlooks an intrinsic property of time series data. Despite being branded as ``direct" forecasting, any time series model naturally possesses an overlooked capacity for temporal extrapolation. By concatenating its own recent predictions with partial historical observation, a model can be perpetually fed back into itself to project into the distant future predictions. Consequently, to see far, expanding the architectural output horizon to match an extreme target horizon is not the sole, nor the optimal, solution for LTSF. Due to the poor performance of ``training far, testing close” and concerns over extrapolation errors, the ``training close, testing far” approach has long been overlooked by the community. The rigid synchronization of output horizon and evaluation horizon is a historical oversight that conflates the most important hyperparameters of model with the task's evaluation metric.

In this paper, we challenge this monolithic cornerstone of LTSF and propose a foundational paradigm shift: \underline{\textbf{E}}volutionary \underline{\textbf{F}}orecasting (\textbf{EF}). EF paradigm explicitly decouples the model's intrinsic output horizon from the task-specific evaluation horizon, liberating output horizon to serve as an essentially optional hyperparameter. Under EF paradigm, the future predictions are generated as a sequence of ``Reasoning Blocks". The final output is seamlessly stitched through the iterative application of a single model. From this unified generative perspective, the traditional DF paradigm is merely a special case where the output and evaluation horizons are rigidly coupled, or equivalently, a specific instance of EF employing a ``Teacher Forcing'' strategy during training. In summary, our contributions set a massive upheaval for LTSF research as follows:

\begin{itemize}
    \item \textbf{A Foundational Paradigm Shift:} We propose Evolutionary Forecasting (EF), a generative paradigm that fundamentally decouples the model's architectural output horizon from the task-specific evaluation horizon. By unearthing the latent, neglected extrapolation capabilities of models in LTSF, we propel the field's transition from passive ``Static Mapping" to autonomous ``Evolutionary Reasoning".

    \item \textbf{``One-for-All'' Generalization \& Extreme Extrapolation:} Extensive experiments demonstrate that under the EF paradigm, a model requires only a single training session to consistently surpass the collective performance of horizon-specific, re-customized, and re-trained DF models across diverse evaluation horizons. Furthermore, in extreme extrapolation scenarios, the EF model exhibits robust asymptotic stability, yielding performance that even exceeds that of DF, thereby effectively dismantling prevailing concerns regarding extrapolation error accumulation.

    \item \textbf{Deconstructing the Optimization Pathology:} We formally identify and visualize the root cause of sub-optimality in the prevailing Direct Forecasting (DF) paradigm. Our analysis reveals a severe Gradient Conflict driven by ``Distal Dominance'', where distant gradients hijack the optimization trajectory, leading to the chronic under-fitting of near-term time steps.
        
\end{itemize}

\begin{figure}[!h]
    \centering
    \includegraphics[width=\linewidth]{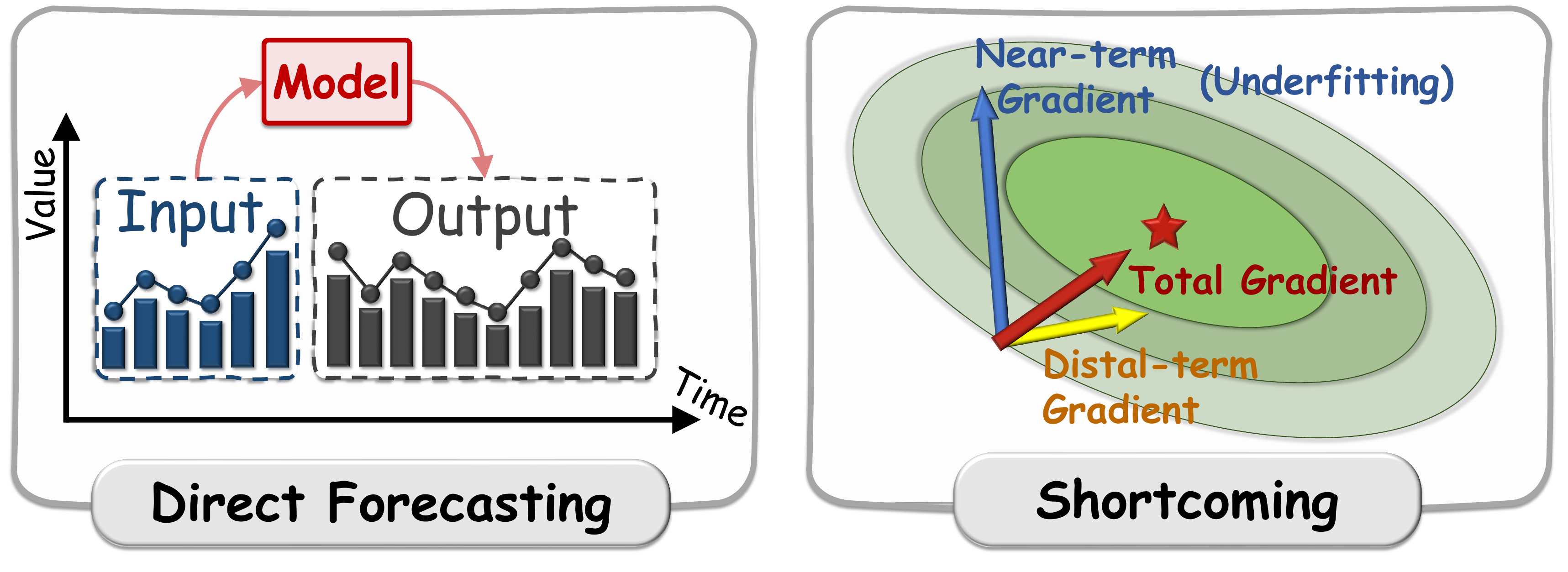}
    \caption{The workflow and shortcoming of Direct Forecasting.}
    \label{fig:DF_workflow}
\end{figure} 

\begin{figure*}[!t]
    \centering
    \includegraphics[width=\linewidth]{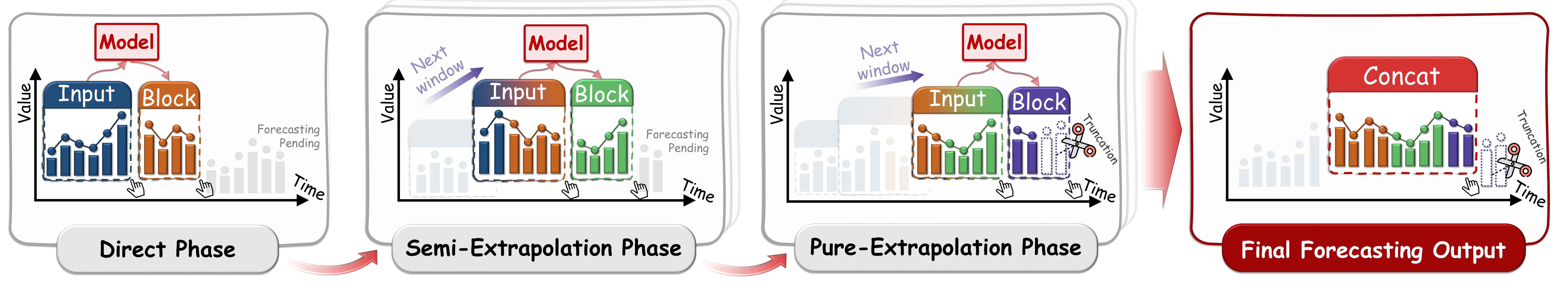}
    \caption{The workflow of Evolutionary Forecasting. \textcolor[HTML]{214D7B}{\textbf{Blue items}} are historical observation.  }
    \label{fig:EF_workflow}
\end{figure*}

\section{Related Work}\label{sec:related}
The paradigm of deep learning-based LTSF models have undergone a shift from Auto Regressive to Direct Forecasting. 

\textbf{Auto Regressive (AR) Paradigm.} Prior to the ascendancy of Transformer~\cite{vaswani2017attention,wen2022transformers}, sequence modeling in LTSF was predominantly governed by the AR paradigm. Ranging from statistical foundations like ARIMA~\cite{box1976analysis,box2015time} to deep learning pioneers such as DeepAR~\cite{salinas2020deepar} and DeepState~\cite{rangapuram2018deep}, these models typically operate on a recursive point-wise mechanism: the model predicts a single future value at a time, feeding this prediction back as input to roll out the subsequent sequence. While this step-by-step generation theoretically allows for infinite horizon forecasting, it inherently suffers from error accumulation~\cite{lim2021time,torres2021deep}. This renders long-term rollouts susceptible to significant distribution shifts and instability, severely limiting their efficacy for extended forecasting horizons.

\textbf{Direct Forecasting (DF) Paradigm.} The advent of Transformer architectures revolutionized sequence modeling by introducing parallel computing mechanisms. Capitalizing on this, Informer~\cite{zhou2021informer} pioneered the transition to the Direct Forecasting (DF) framework in LTSF. This breakthrough catalyzed a proliferation of Transformer-based variants under DF paradigm~\cite{wu2021autoformer,zhou2022fedformer,liu2022pyraformer,zhang2023crossformer}. However, subsequent research, exemplified by DLinear~\cite{zeng2023transformers}, challenged the efficacy of complex attention mechanisms for continuous time series data, citing issues with overfitting and distribution shift. This prompted a development towards lightweight MLP and even Linear-based models~\cite{xu2023fits,lin2024sparsetsf,huangtimebase}. However, these architectures are structurally devoid of inherent extrapolation capabilities and are strictly bound to their training length as fixed-size mappings. Consequently, this limitation exacerbates the rigid coupling between the output horizon and the inference evaluation horizon . This constraint means the model's predictive capability is ``locked'' during training, necessitating the re-training of separate models for every varying evaluation horizon—a computationally prohibitive protocol that our proposed EF paradigm aims to resolve by decoupling training from inference. More discussion is in Appendix~\ref{app:relaterwork}.

\section{Preliminaries}\label{sec:pre}
The common objective of LTSF is to train a model $\mathcal{F}_\theta: \mathbb{R}^{T \times C} \to \mathbb{R}^{L \times C}$ that maps the historical observations $\mathbf{X} = [\mathbf{x}_{1}, \mathbf{x}_{2}, \dots, \mathbf{x}_{T}] \in \mathbb{R}^{T \times C}$ into $\hat{\mathbf{Y}} = \mathcal{F}_\theta(\mathbf{X})$ as the future prediction of $\mathbf{Y} = [\mathbf{x}_{T+1}, \mathbf{x}_{T+2}, \dots, \mathbf{x}_{T+L}] \in \mathbb{R}^{L \times C}$, where $\mathbf{x}_t \in \mathbb{R}^{C}$ denotes the $C$-dimensional observation at time step $t$, which can represent multiple features of a single entity or synchronous observations across $C$ different entities. $T\in\mathbb{Z}$ is the input length and $L\in\mathbb{Z}$ is the specific output horizon for forecasting. The current mainstream paradigm for LTSF is Direct Forecasting (as shown in the left of Figure~\ref{fig:DF_workflow}), where the model's output horizon is equivalent to the evaluation horizon $H\in\mathbb{Z}$, and the predictions is fully obtained through a single forward process.

\section{A Unifying View of LTSF}
In this section, we formalize a generalized paradigm for LTSF tasks through a high-level abstraction termed \underline{\textbf{E}}volutionary \underline{\textbf{F}}orecasting (EF). EF effectively decouples the evaluation horizon $H$ from the output horizon $L$, thereby uncovering a previously overlooked dimension of freedom in model design and optimization. The model $\mathcal{F}_\theta$ is treated as a recurrent operator that iteratively generates \textbf{Reasoning Blocks} of length $L$, which are then concatenated to consist an arbitrary evaluation horizon $H$, where $H$ is typically a multiple of $L$ (i.e., $H = \lfloor K \cdot L\rfloor$). This formalization allows us to investigate the intrinsic dynamical properties of the model independently of the specific task requirements. The workflow of EF is in Figure~\ref{fig:EF_workflow} and Algorithm~\ref{alg:ef_framework}.

Specifically, we first provide a rigorous mathematical definition of the EF paradigm in Section~\ref{sec:defbar}. Then we demonstrate its unifying nature by showing that current mainstream DF paradigm is merely a degenerate special cases within EF paradigm in Section~\ref{sec:degexamples}. 

\subsection{Description of Evolutionary Forecasting}\label{sec:defbar}
We formally unify LTSF task under EF paradigm. Unlike conventional DF that treat the output horizon as a fixed task constraint like $L=H$, EF treats it as a flexible model hyperparameter. Let $\mathcal{F}_\theta: \mathbb{R}^{T \times C} \to \mathbb{R}^{L \times C}$ be the model of output horizon $L$, we can define that as follows,
\begin{definition}\label{def:barltsf}\textbf{(Evolutionary Forecasting (EF))}. 
     Given a target task with an evaluation horizon $H$, the final prediction output $\hat{\mathbf{Y}} \in \mathbb{R}^{H \times C}$ is reasoned through $K = \lceil H/L \rceil$ steps iterative applications of $\mathcal{F}_\theta$ as follows:
    \begin{equation}
      \hat{\mathbf{Y}} = \left[ \mathbf{B}^{(1)}\mid \mathbf{B}^{(2)}\mid \dots\mid \mathbf{B}^{(K)} \right]_{1:H},
    \end{equation}
    where $\mathbf{B}^{(k)} \in \mathbb{R}^{L \times C}$ is the $k$-th Reasoning Block, and the output $\hat{\mathbf{Y}}$ is the concatenation of all blocks along the temporal dimension, truncated to evaluation horizon $H$.
\end{definition}

\begin{definition}\label{def:barltsf}\textbf{(The $k$-th Reasoning Block ($k$-Block))}. The $k$-th Reasoning Block $\mathbf{B}^{(k)}$ is the $k$-th step output of EF as $\mathbf{B}^{(k)}=\mathcal{F}_{\theta}(\tilde{\mathbf{X}}^{(k)})\in\mathbb{R}^{L\times C}$ where the input of $k$-th step $\tilde{\mathbf{X}}^{(k)} \in \mathbb{R}^{T \times C}$ for this block is rigorously defined as:
\begin{equation*}
  \tilde{\mathbf{X}}^{(k)}=\begin{cases}
  \;\mathbf{X}, & \text{if }k=1, \\
    [\mathbf{X}_{(k-1)L:T},\hat{\textbf{Y}}_{1:(k-1)L}], & \text{if }1<k<\frac{T}{L}+1, \\
    \;\hat{\textbf{Y}}_{(k-1)L-T:(k-1)L}, & \text{if }k\geq \frac{T}{L}+1.
  \end{cases}
\end{equation*}
\end{definition}
Concretely, the input of $k$-Block, $\tilde{\mathbf{X}}^{(k)}$, transitions through three distinct phases based on the relationship between the cumulative predicted length $(k-1)L$ and the original input length $T$: \ding{202} \textbf{Direct Phase}: when $k=1$, the input data is exactly the complete historic observation $\mathbf{X}\in\mathbb{R}^{T\times C}$. This phase can be seen as the start-step of the EF. \ding{203} \textbf{Semi-Extrapolation Phase}: When $0<(k-1)L<T$, the input data is the mixed concatenate of latter portion of historical observation $\mathbf{X}_{(k-1)L:T}$ and the current cumulative predicted output $\hat{\textbf{Y}}_{1:(k-1)L}=\left[ \mathbf{B}^{(1)}\mid \mathbf{B}^{(2)}\mid \dots\mid \mathbf{B}^{(k-1)} \right]$ since the input length of model is fixed. This phase represents a Mixed-Context state, where the model $\mathcal{F}_\theta$ tends to bridge the empirical historical observation with its own generated predictions, effectively smoothing the transition to extrapolation. \ding{204} \textbf{Pure-Extrapolation Phase}: When $(k-1)L \geq T$, the input is entirely composed of previously generated predictions $\hat{\textbf{Y}}_{(k-1)L-T:(k-1)L}$. In this phase, the logic consistency of the model $\mathcal{F}_\theta$ is the sole determinant of long-term stability. The corresponding visualized workflow of EF is in Figure~\ref{fig:EF_workflow} and  Algorithm~\ref{alg:ef_framework}. 

\begin{remark}\label{rem:df_special_case}
(Output Horizon Surplus). When $L > H$, the model similarly operates in a single step ($K=1$) of Direct Phase, and the final output is obtained by truncating the first Reasoning Block: $\hat{\mathbf{Y}} = \mathbf{B}^{(1)}_{1:H} = \mathcal{F}_{\theta}(\mathbf{X})_{1:H}$. 
\end{remark}

\subsection{Direct Forecasting as a Special Case of EF}\label{sec:degexamples}

To establish the foundational nature of EF paradigm, we demonstrate that the ubiquitous DF paradigm is just a restricted, single-step instantiation as a specific boundary case in EF. This relationship can be analyzed through two distinct perspectives:

\textbf{View I: The Degeneration of Output Horizon ($L= H$).} From the perspective of task decomposition, DF can be viewed as an EF configuration where the output horizon $L$ is forced to synchronize with the evaluation horizon $H$ under the condition $K=1$. The iterative reasoning chain collapses into a single-step mapping:
\begin{equation}
\hat{\mathbf{Y}} = \mathcal{F}_\theta(\mathbf{X})_{1:H}, \quad \text{where } L \geq H.
\end{equation}
In this case, the model is deprived of its reasoning capacity, being forced to forecasting all future time series values in a monolithic projection. Our EF paradigm thus subsumes DF as its most basic, none-iteration variant.

\textbf{View II: The Teacher Forcing Limit of Training.} Even when considering the training mechanism, DF essentially operates as a restricted instance of EF employing a pure Teacher Forcing strategy. In our EF paradigm, the model is trained to progressively predict future blocks, ideally learning to handle inputs derived from its own previous outputs. Conversely, the traditional DF training protocol strictly constructs input windows using only the historical ground truth $\mathbf{X}$ and $\mathbf{Y}$, completely bypassing the extrapolation phase. Mathematically, this corresponds to forcing the input of every reasoning block $\tilde{\mathbf{X}}^{(k)}$ to be composed exclusively of ground truth data:
\begin{equation}
\tilde{\mathbf{X}}^{(k)} \equiv [\mathbf{X}_{(k-1)L:T},\textbf{Y}_{1:(k-1)L}], \quad \forall k \in (1, K].
\end{equation}
Under this view, DF is equivalent to an EF model that is perpetually confined to the ``Direct Phase" during training.

\section{Experiments}\label{sec:exp}

In this section, we conduct a comprehensive empirical evaluation to challenge the long-standing, yet previously unquestioned convention in LTSF: the rigid alignment of the model’s output horizon $L$ with the evaluation horizon $H$. We demonstrate that this implicit ``synchronization" is fundamentally sub-optimal, as it inevitably induces gradient conflict—an optimization pathology driven by severe directional conflicts between near-term and distal time steps. 

Our experiments systematically analyse the performance landscape across the $(L, H)$ horizon space under various input length $T$. Rather than viewing horizon selection as a mere byproduct of task requirements, we reveal it to be an intrinsic property of the underlying data dynamics as the essential hyperparameters. Specifically, we focus on the following research questions to validate our paradigm shift:

\begin{itemize}
    \item \textbf{RQ1 (The Decoupling Hypothesis):} Does decoupling the model’s intrinsic output horizon $L$ from the task-specific evaluation horizon $H$ within the EF paradigm consistently yield superior predictive fidelity compared to conventional DF ($L=H$)?
    \item \textbf{RQ2 (``One-for-All'' Generalization):} Can a singular, well-calibrated model under EF by optimal $T^*$ and $L^*$ robustly generalize across all the evaluation horizons, outperforming an ensemble of computationally horizon-specific re-customized and retrained DF?
    \item \textbf{RQ3 (Extreme Extrapolation Stability):} Since the EF paradigm enables unbounded reasoning via the Pure-Extrapolation Phase, can a model optimized for a short output horizon ($L \ll H$) consistently outperform a specialized DF paradigm ($L = H$) on extreme-scale evaluation horizons?
    \item \textbf{RQ4 (The Optimization Pathology):} Fundamentally, what drives the sub-optimal performance of DF that sets $L=H$?
\end{itemize}

\subsection{Experimental Setup}\label{sec:setup}

\textbf{Datasets.} To empirically validate the proposed EF paradigm and the associated phenomena, we conduct extensive experiments on six authoritative real-world datasets spanning diverse domains: ETT series, Electricity, Weather, Traffic, Exchange and ILI. These datasets represent a wide spectrum of temporal dynamics, ranging from strong periodicity to complex stochastic. Detailed statistics for each dataset are provided in Table \ref{tab:dataset}. Following standard protocols \cite{wu2021autoformer, wu2022timesnet, qiu2024tfb}, we adopt a chronological split ratio of 6:2:2 for the ETT series and 7:1:2 for all other datasets.

\textbf{Configurations.} Aligning with our EF formalization, the search space for three temporal scales: the input length $T$, the output horizon $L$, and the evaluation horizon $H$ strictly adhere to established benchmarks \cite{wu2021autoformer, wu2022timesnet, qiu2024tfb}. For the ILI datasets, $T, L, H \in \{24, 36, 48, 60\}$. For the remaining six datasets, we explore the combinations of $T, L, H \in \{96, 192, 336, 720\}$. All models are optimized using the ADAM optimizer \cite{kingma2014adam}. We employ Mean Squared Error (MSE) as the loss function. For each experiment, the training process was conducted for a maximum of 100 epochs with an early stopping mechanism with a max patience of 10 epochs, monitoring the validation loss for sufficient training.

\begin{table}[htbp]
  \centering
  \aboverulesep=0pt
  \belowrulesep=0pt
  \setlength{\arrayrulewidth}{1.0pt}
  \caption{Statistical summary and experimental configurations of the datasets. ``Split” refers to the chronological partition ratio for training, validation, and testing. ``Window Options” denotes the candidate space for $T,L,H$.}
  \resizebox{\linewidth}{!}{
    \begin{tabular}{lllc|cc}
    \toprule
    \rowcolor[rgb]{ .867,  .875,  .91}\textbf{Datasets} & \textbf{Domain} & \textbf{Samples} & \textbf{\# Channels} & \textbf{Split} & \textbf{Window Options} \\
    \midrule
    ETTh1 & Electricity & 1 hour & 7     & 6:2:2 & \{96, 192, 336, 720\} \\
    \rowcolor[rgb]{ .949,  .949,  .949}ETTh2 & Electricity & 1 hour & 7     & 6:2:2 & \{96, 192, 336, 720\} \\
    Exchange & Economic & 1 day & 8     & 7:1:2 & \{96, 192, 336, 720\} \\
    \rowcolor[rgb]{ .949,  .949,  .949}Weather & Environment & 10 mins & 21    & 7:1:2 & \{96, 192, 336, 720\} \\
    Traffic & Traffic & 1 hour & 862   & 7:1:2 & \{96, 192, 336, 720\} \\
    \rowcolor[rgb]{ .949,  .949,  .949}ILI   & Health & 1 week & 7     & 7:1:2 & \{24, 36, 48, 60\} \\
    \bottomrule
    \end{tabular}}%
  \label{tab:dataset}%
\end{table}%

\textbf{Time series models.} To evaluate the cross-architecture universality of the EF paradigm and since models of DF have become the mainstream and leading approach for deep learning-based LTSF, we select a diverse suite of eight state-of-the-art deep learning models with DF as following: DLinear~\cite{zeng2023transformers}, PatchTST~\cite{nie2023time}, iTransformer~\cite{liu2023itransformer}, TimeMixer~\cite{wang2024timemixer}, SparseTSF~\cite{lin2024sparsetsf}, TimeBase~\cite{huangtimebase}, TimeBridge~\cite{liu2024timebridge} and TimeEmb~\cite{xia2025timeemb}. Each model with different fixed input length $T$ or output horizon $L$ is treated as an independent training subject. To ensure a rigorous and unbiased comparison, we strictly follow the official implementations provided by the authors or the optimal hyper-parameters established in the official repository or reputable benchmarks \cite{qiu2024tfb, wu2022timesnet}, thereby ensuring that the observed gains are intrinsic to the EF paradigm. 

\textbf{Experimental environment.} Our code framework is modified from Time Series Library\footnote{https://github.com/thuml/Time-Series-Library} on Python 3.12.3 and PyTorch 2.5.1. All experiments are performed on a server equipped with an Intel(R) Xeon(R) Platinum 8474C CPU and NVIDIA GeForce RTX 4090D GPUs with 24 GB. We utilized CUDA 12.4 and cuDNN 9.0.1 for GPU.

\subsection{Main Experiments: Paradigm Shift (RQ1 \& RQ2)}\label{sec:mainexp}

In this section, we provide a comprehensive empirical evaluation to demonstrate the superiority of the EF paradigm over the conventional DF paradigm. Given the expansive $(T, L, H)$ configuration space—comprising $4^3$ possible combinations for each model-dataset pair—we categorize our results into two complementary perspectives to maintain clarity within the page limit: \ding{202} \textbf{Granular Numerical Comparison}: Table~\ref{tab:main} reports detailed performance across the $(L, H)$ space with a fixed input length $T=720$ for ETTh1 (strong seasonality) and Weather (Weak seasonality). Performance is quantified by Mean Squared Error (MSE) and Mean Absolute Error (MAE). \ding{203} \textbf{Global Statistical Comparison}: Figure~\ref{fig:placeholder} visualizes the Win Ratio, which is the normalized aggregate Win Count of all comparison objects across all baseline models and evaluation horizons $H$. Comprehensive numerical results for all configurations are provided in Tables~\ref{tab:dlinear_all} to \ref{tab:timeemb_all} in Appendix~\ref{app:mainexp}.

\begin{table*}[!t]
  \centering
  \aboverulesep=0pt
  \belowrulesep=0pt
  \setlength{\arrayrulewidth}{1.0pt}
  \setlength{\tabcolsep}{2pt}
  \caption{Performance landscape of ETTh1 and Weather datasets on the $(L, H)$ horizon space. Results under the conventional DF paradigm ($L=H$) are highlighted in \colorbox[HTML]{F6E9C5}{Orange Background}. The best performance for a fixed $H$ is denoted in \textcolor{red}{\textbf{Bold}}. $\Delta$ is the improvement ($\%$ ) from the best results in to DF's results. The positive improvements are in \textcolor{blue}{Blue}.}
  \resizebox{\linewidth}{!}{
    \begin{tabular}{cc|cc|cc|cc|cc|c|cc|cc|cc|cc}
    \toprule
    \multicolumn{2}{c|}{\textbf{Datasets}} & \multicolumn{8}{c}{\textbf{ETTh1 ($T=720$)}}                    & \multicolumn{1}{c}{} & \multicolumn{8}{c}{\textbf{Weather ($T=720$)}} \\
    \midrule
    \multirow{2}[0]{*}{\textbf{Models}} & \multirow{2}[0]{*}{\diagbox{$L$}{$H$}} & \multicolumn{2}{c|}{{$96$}} & \multicolumn{2}{c|}{{$192$}} & \multicolumn{2}{c|}{{$336$}} & \multicolumn{2}{c|}{{$720$}} & \multirow{2}[0]{*}{\diagbox{$L$}{$H$}} & \multicolumn{2}{c|}{{$96$}} & \multicolumn{2}{c|}{{$192$}} & \multicolumn{2}{c|}{{$336$}} & \multicolumn{2}{c}{{$720$}} \\
          &       & {MSE} & {MAE} & {MSE} & {MAE} & {MSE} & {MAE} & {MSE} & {MAE} &       & {MSE} & {MAE} & {MSE} & {MAE} & {MSE} & {MAE} & {MSE} & {MAE} \\
    \midrule
    \midrule
    \multirow{5}[0]{*}{\textbf{DLinear}} & {$96$} & \cellcolor[HTML]{F6E9C5}0.404 & \cellcolor[HTML]{F6E9C5}0.421 & 0.433 & 0.442 & 0.481 & 0.499 & 0.481 & 0.499 & {$96$} & \cellcolor[HTML]{F6E9C5}\textcolor[rgb]{ 1,  0,  0}{\textbf{0.168}} & \cellcolor[HTML]{F6E9C5}\textcolor[rgb]{ 1,  0,  0}{\textbf{0.227}} & \textcolor[rgb]{ 1,  0,  0}{\textbf{0.211}} & \textcolor[rgb]{ 1,  0,  0}{\textbf{0.266}} & \textcolor[rgb]{ 1,  0,  0}{\textbf{0.255}} & \textcolor[rgb]{ 1,  0,  0}{\textbf{0.304}} & 0.317 & \textcolor[rgb]{ 1,  0,  0}{\textbf{0.356}} \\
          & {$192$} & \textcolor[rgb]{ 1,  0,  0}{\textbf{0.369}} & \textcolor[rgb]{ 1,  0,  0}{\textbf{0.399}} & \cellcolor[HTML]{F6E9C5}\textcolor[rgb]{ 1,  0,  0}{\textbf{0.405}} & \cellcolor[HTML]{F6E9C5}\textcolor[rgb]{ 1,  0,  0}{\textbf{0.422}} & \textcolor[rgb]{ 1,  0,  0}{\textbf{0.434}} & \textcolor[rgb]{ 1,  0,  0}{\textbf{0.443}} & \textcolor[rgb]{ 1,  0,  0}{\textbf{0.480}} & \textcolor[rgb]{ 1,  0,  0}{\textbf{0.498}} & {$192$} & 0.169 & 0.228 & \cellcolor[HTML]{F6E9C5}\textcolor[rgb]{ 1,  0,  0}{\textbf{0.211}} & \cellcolor[HTML]{F6E9C5}0.267 & 0.256 & 0.305 & 0.317 & 0.357 \\
          & {$336$} & 0.380 & 0.412 & 0.410 & 0.429 & \cellcolor[HTML]{F6E9C5}0.440 & \cellcolor[HTML]{F6E9C5}0.452 & 0.496 & 0.512 & {$336$} & \textcolor[rgb]{ 1,  0,  0}{\textbf{0.168}} & 0.228 & \textcolor[rgb]{ 1,  0,  0}{\textbf{0.211}} & 0.267 & \cellcolor[HTML]{F6E9C5}0.256 & \cellcolor[HTML]{F6E9C5}0.305 & \textcolor[rgb]{ 1,  0,  0}{\textbf{0.316}} & \textcolor[rgb]{ 1,  0,  0}{\textbf{0.356}} \\
          & {$720$} & 0.411 & 0.431 & 0.439 & 0.452 & 0.488 & 0.507 & \cellcolor[HTML]{F6E9C5}0.488 & \cellcolor[HTML]{F6E9C5}0.507 & {$720$} & 0.169 & 0.230 & 0.212 & 0.270 & 0.257 & 0.308 & \cellcolor[HTML]{F6E9C5}\textcolor[rgb]{ 1,  0,  0}{\textbf{0.316}} & \cellcolor[HTML]{F6E9C5}0.357 \\
\cmidrule{2-19}          & \textbf{$\Delta$} & \textcolor{blue}{{8.66\%}} & \textcolor{blue}{{5.23\%}} & \textcolor{black}{{0.00\%}} & \textcolor{black}{{0.00\%}} & \textcolor{blue}{{1.36\%}} & \textcolor{blue}{{1.99\%}} & \textcolor{blue}{{1.64\%}} & \textcolor{blue}{{1.78\%}} & \textbf{$\Delta$} & \textcolor{black}{{0.00\%}} & \textcolor{black}{{0.00\%}} & \textcolor{black}{{0.00\%}} & \textcolor{blue}{{0.37\%}} & \textcolor{blue}{{0.39\%}} & \textcolor{blue}{{0.33\%}} & \textcolor{black}{{0.00\%}} & \textcolor{blue}{{0.28\%}} \\
    \midrule
    \midrule
    \multirow{5}[0]{*}{\textbf{PatchTST}} & {$96$} & \cellcolor[HTML]{F6E9C5}\textcolor[rgb]{ 1,  0,  0}{\textbf{0.405}} & \cellcolor[HTML]{F6E9C5}\textcolor[rgb]{ 1,  0,  0}{\textbf{0.424}} & \textcolor[rgb]{ 1,  0,  0}{\textbf{0.436}} & \textcolor[rgb]{ 1,  0,  0}{\textbf{0.445}} & \textcolor[rgb]{ 1,  0,  0}{\textbf{0.450}} & \textcolor[rgb]{ 1,  0,  0}{\textbf{0.457}} & \textcolor[rgb]{ 1,  0,  0}{\textbf{0.455}} & \textcolor[rgb]{ 1,  0,  0}{\textbf{0.473}} & {$96$} & \cellcolor[HTML]{F6E9C5}\textcolor[rgb]{ 1,  0,  0}{\textbf{0.150}} & \cellcolor[HTML]{F6E9C5}\textcolor[rgb]{ 1,  0,  0}{\textbf{0.202}} & \textcolor[rgb]{ 1,  0,  0}{\textbf{0.198}} & \textcolor[rgb]{ 1,  0,  0}{\textbf{0.249}} & \textcolor[rgb]{ 1,  0,  0}{\textbf{0.255}} & \textcolor[rgb]{ 1,  0,  0}{\textbf{0.293}} & 0.345 & 0.354 \\
          & {$192$} & 0.418 & 0.437 & \cellcolor[HTML]{F6E9C5}0.455 & \cellcolor[HTML]{F6E9C5}0.460 & 0.469 & 0.469 & 0.473 & 0.481 & {$192$} & 0.159 & 0.215 & \cellcolor[HTML]{F6E9C5}0.204 & \cellcolor[HTML]{F6E9C5}0.255 & 0.257 & 0.295 & \textcolor[rgb]{ 1,  0,  0}{\textbf{0.331}} & \textcolor[rgb]{ 1,  0,  0}{\textbf{0.345}} \\
          & {$336$} & 0.437 & 0.443 & 0.500 & 0.487 & \cellcolor[HTML]{F6E9C5}0.503 & \cellcolor[HTML]{F6E9C5}0.497 & 0.514 & 0.509 & {$336$} & 0.166 & 0.227 & 0.214 & 0.267 & \cellcolor[HTML]{F6E9C5}0.266 & \cellcolor[HTML]{F6E9C5}0.306 & 0.334 & 0.352 \\
          & {$720$} & 0.479 & 0.479 & 0.501 & 0.491 & 0.521 & 0.502 & \cellcolor[HTML]{F6E9C5}0.521 & \cellcolor[HTML]{F6E9C5}0.514 & {$720$} & 0.167 & 0.224 & 0.213 & 0.264 & 0.269 & 0.304 & \cellcolor[HTML]{F6E9C5}0.340 & \cellcolor[HTML]{F6E9C5}0.352 \\
\cmidrule{2-19}          & \textbf{$\Delta$} & \textcolor{black}{{0.00\%}} & \textcolor{black}{{0.00\%}} & \textcolor{blue}{{4.18\%}} & \textcolor{blue}{{3.26\%}} & \textcolor{blue}{{10.54\%}} & \textcolor{blue}{{8.05\%}} & \textcolor{blue}{{12.67\%}} & \textcolor{blue}{{7.98\%}} & \textbf{$\Delta$} & \textcolor{black}{{0.00\%}} & \textcolor{black}{{0.00\%}} & \textcolor{blue}{{2.94\%}} & \textcolor{blue}{{2.35\%}} & \textcolor{blue}{{4.14\%}} & \textcolor{blue}{{4.25\%}} & \textcolor{blue}{{2.65\%}} & \textcolor{blue}{{1.99\%}} \\
    \midrule
    \midrule
    \multirow{5}[0]{*}{\textbf{iTransformer}} & {$96$} & \cellcolor[HTML]{F6E9C5}\textcolor[rgb]{ 1,  0,  0}{\textbf{0.392}} & \cellcolor[HTML]{F6E9C5}\textcolor[rgb]{ 1,  0,  0}{\textbf{0.424}} & \textcolor[rgb]{ 1,  0,  0}{\textbf{0.425}} & \textcolor[rgb]{ 1,  0,  0}{\textbf{0.444}} & \textcolor[rgb]{ 1,  0,  0}{\textbf{0.449}} & \textcolor[rgb]{ 1,  0,  0}{\textbf{0.462}} & \textcolor[rgb]{ 1,  0,  0}{\textbf{0.476}} & \textcolor[rgb]{ 1,  0,  0}{\textbf{0.493}} & {$96$} & \cellcolor[HTML]{F6E9C5}0.174 & \cellcolor[HTML]{F6E9C5}0.228 & 0.226 & 0.271 & 0.284 & 0.310 & 0.382 & 0.368 \\
          & {$192$} & 0.402 & 0.433 & \cellcolor[HTML]{F6E9C5}0.437 & \cellcolor[HTML]{F6E9C5}0.455 & 0.467 & 0.477 & 0.513 & 0.520 & {$192$} & \textcolor[rgb]{ 1,  0,  0}{\textbf{0.166}} & \textcolor[rgb]{ 1,  0,  0}{\textbf{0.219}} & \cellcolor[HTML]{F6E9C5}\textcolor[rgb]{ 1,  0,  0}{\textbf{0.222}} & \cellcolor[HTML]{F6E9C5}\textcolor[rgb]{ 1,  0,  0}{\textbf{0.264}} & \textcolor[rgb]{ 1,  0,  0}{\textbf{0.278}} & \textcolor[rgb]{ 1,  0,  0}{\textbf{0.302}} & \textcolor[rgb]{ 1,  0,  0}{\textbf{0.343}} & \textcolor[rgb]{ 1,  0,  0}{\textbf{0.345}} \\
          & {$336$} & 0.401 & 0.433 & 0.431 & 0.451 & \cellcolor[HTML]{F6E9C5}0.457 & \cellcolor[HTML]{F6E9C5}0.470 & 0.494 & 0.508 & {$336$} & 0.170 & 0.225 & \textcolor[rgb]{ 1,  0,  0}{\textbf{0.222}} & 0.268 & \cellcolor[HTML]{F6E9C5}0.280 & \cellcolor[HTML]{F6E9C5}0.309 & 0.354 & 0.357 \\
          & {$720$} & 0.473 & 0.477 & 0.502 & 0.494 & 0.517 & 0.505 & \cellcolor[HTML]{F6E9C5}0.553 & \cellcolor[HTML]{F6E9C5}0.537 & {$720$} & 0.177 & 0.234 & 0.226 & 0.273 & 0.281 & 0.311 & \cellcolor[HTML]{F6E9C5}0.352 & \cellcolor[HTML]{F6E9C5}0.361 \\
\cmidrule{2-19}          & \textbf{$\Delta$} & \textcolor{black}{{0.00\%}} & \textcolor{black}{{0.00\%}} & \textcolor{blue}{{2.75\%}} & \textcolor{blue}{{2.42\%}} & \textcolor{blue}{{1.75\%}} & \textcolor{blue}{{1.70\%}} & \textcolor{blue}{{13.92\%}} & \textcolor{blue}{{8.19\%}} & \textbf{$\Delta$} & \textcolor{blue}{{4.60\%}} & \textcolor{blue}{{3.95\%}} & \textcolor{black}{{0.00\%}} & \textcolor{black}{{0.00\%}} & \textcolor{blue}{{0.71\%}} & \textcolor{blue}{{2.27\%}} & \textcolor{blue}{{2.56\%}} & \textcolor{blue}{{4.43\%}} \\
    \midrule
    \midrule
    \multirow{5}[0]{*}{\textbf{TimeMixer}} & {$96$} & \cellcolor[HTML]{F6E9C5}\textcolor[rgb]{ 1,  0,  0}{\textbf{0.384}} & \cellcolor[HTML]{F6E9C5}\textcolor[rgb]{ 1,  0,  0}{\textbf{0.414}} & \textcolor[rgb]{ 1,  0,  0}{\textbf{0.420}} & \textcolor[rgb]{ 1,  0,  0}{\textbf{0.437}} & \textcolor[rgb]{ 1,  0,  0}{\textbf{0.443}} & \textcolor[rgb]{ 1,  0,  0}{\textbf{0.452}} & \textcolor[rgb]{ 1,  0,  0}{\textbf{0.489}} & \textcolor[rgb]{ 1,  0,  0}{\textbf{0.495}} & {$96$} & \cellcolor[HTML]{F6E9C5}0.154 & \cellcolor[HTML]{F6E9C5}0.209 & 0.207 & 0.255 & 0.269 & 0.298 & 0.334 & 0.342 \\
          & {$192$} & 0.400 & 0.426 & \cellcolor[HTML]{F6E9C5}0.443 & \cellcolor[HTML]{F6E9C5}0.454 & 0.470 & 0.472 & 0.518 & 0.515 & {$192$} & 0.157 & 0.213 & \cellcolor[HTML]{F6E9C5}0.199 & \cellcolor[HTML]{F6E9C5}0.250 & 0.250 & 0.289 & 0.320 & 0.338 \\
          & {$336$} & 0.412 & 0.437 & 0.460 & 0.467 & \cellcolor[HTML]{F6E9C5}0.494 & \cellcolor[HTML]{F6E9C5}0.490 & 0.534 & 0.523 & {$336$} & \textcolor[rgb]{ 1,  0,  0}{\textbf{0.150}} & \textcolor[rgb]{ 1,  0,  0}{\textbf{0.206}} & \textcolor[rgb]{ 1,  0,  0}{\textbf{0.196}} & \textcolor[rgb]{ 1,  0,  0}{\textbf{0.248}} & \cellcolor[HTML]{F6E9C5}\textcolor[rgb]{ 1,  0,  0}{\textbf{0.245}} & \cellcolor[HTML]{F6E9C5}0.286 & \textcolor[rgb]{ 1,  0,  0}{\textbf{0.312}} & 0.332 \\
          & {$720$} & 0.412 & 0.431 & 0.454 & 0.455 & 0.488 & 0.474 & \cellcolor[HTML]{F6E9C5}0.548 & \cellcolor[HTML]{F6E9C5}0.522 & {$720$} & 0.158 & 0.212 & 0.201 & 0.250 & 0.249 & \textcolor[rgb]{ 1,  0,  0}{\textbf{0.285}} & \cellcolor[HTML]{F6E9C5}0.313 & \cellcolor[HTML]{F6E9C5}\textcolor[rgb]{ 1,  0,  0}{\textbf{0.331}} \\
\cmidrule{2-19}          & \textbf{$\Delta$} & \textcolor{black}{{0.00\%}} & \textcolor{black}{{0.00\%}} & \textcolor{blue}{{5.19\%}} & \textcolor{blue}{{3.74\%}} & \textcolor{blue}{{10.32\%}} & \textcolor{blue}{{7.76\%}} & \textcolor{blue}{{10.77\%}} & \textcolor{blue}{{5.17\%}} & \textbf{$\Delta$} & \textcolor{blue}{{2.60\%}} & \textcolor{blue}{{1.44\%}} & \textcolor{blue}{{1.51\%}} & \textcolor{blue}{{0.80\%}} & \textcolor{black}{{0.00\%}} & \textcolor{blue}{{0.35\%}} & \textcolor{blue}{{0.32\%}} & \textcolor{black}{{0.00\%}} \\
    \midrule
    \midrule
    \multirow{5}[0]{*}{\textbf{SparseTSF}} & {$96$} & \cellcolor[HTML]{F6E9C5}\textcolor[rgb]{ 1,  0,  0}{\textbf{0.450}} & \cellcolor[HTML]{F6E9C5}\textcolor[rgb]{ 1,  0,  0}{\textbf{0.457}} & \textcolor[rgb]{ 1,  0,  0}{\textbf{0.472}} & \textcolor[rgb]{ 1,  0,  0}{\textbf{0.471}} & 0.483 & \textcolor[rgb]{ 1,  0,  0}{\textbf{0.479}} & \textcolor[rgb]{ 1,  0,  0}{\textbf{0.479}} & \textcolor[rgb]{ 1,  0,  0}{\textbf{0.491}} & {$96$} & \cellcolor[HTML]{F6E9C5}\textcolor[rgb]{ 1,  0,  0}{\textbf{0.170}} & \cellcolor[HTML]{F6E9C5}\textcolor[rgb]{ 1,  0,  0}{\textbf{0.224}} & \textcolor[rgb]{ 1,  0,  0}{\textbf{0.213}} & \textcolor[rgb]{ 1,  0,  0}{\textbf{0.260}} & 0.259 & \textcolor[rgb]{ 1,  0,  0}{\textbf{0.294}} & 0.325 & 0.340 \\
          & {$192$} & \textcolor[rgb]{ 1,  0,  0}{\textbf{0.450}} & 0.458 & \cellcolor[HTML]{F6E9C5}0.473 & \cellcolor[HTML]{F6E9C5}0.472 & \textcolor[rgb]{ 1,  0,  0}{\textbf{0.481}} & \textcolor[rgb]{ 1,  0,  0}{\textbf{0.479}} & 0.480 & 0.492 & {$192$} & \textcolor[rgb]{ 1,  0,  0}{\textbf{0.170}} & \textcolor[rgb]{ 1,  0,  0}{\textbf{0.224}} & \cellcolor[HTML]{F6E9C5}\textcolor[rgb]{ 1,  0,  0}{\textbf{0.213}} & \cellcolor[HTML]{F6E9C5}\textcolor[rgb]{ 1,  0,  0}{\textbf{0.260}} & \textcolor[rgb]{ 1,  0,  0}{\textbf{0.258}} & \textcolor[rgb]{ 1,  0,  0}{\textbf{0.294}} & 0.323 & 0.340 \\
          & {$336$} & 0.452 & 0.458 & 0.474 & 0.472 & \cellcolor[HTML]{F6E9C5}0.487 & \cellcolor[HTML]{F6E9C5}0.482 & 0.483 & 0.494 & {$336$} & \textcolor[rgb]{ 1,  0,  0}{\textbf{0.170}} & 0.225 & \textcolor[rgb]{ 1,  0,  0}{\textbf{0.213}} & \textcolor[rgb]{ 1,  0,  0}{\textbf{0.260}} & \cellcolor[HTML]{F6E9C5}\textcolor[rgb]{ 1,  0,  0}{\textbf{0.258}} & \cellcolor[HTML]{F6E9C5}0.295 & 0.322 & \textcolor[rgb]{ 1,  0,  0}{\textbf{0.339}} \\
          & {$720$} & 0.454 & 0.460 & 0.475 & 0.473 & 0.487 & 0.482 & \cellcolor[HTML]{F6E9C5}0.489 & \cellcolor[HTML]{F6E9C5}0.498 & {$720$} & \textcolor[rgb]{ 1,  0,  0}{\textbf{0.170}} & \textcolor[rgb]{ 1,  0,  0}{\textbf{0.224}} & \textcolor[rgb]{ 1,  0,  0}{\textbf{0.213}} & \textcolor[rgb]{ 1,  0,  0}{\textbf{0.260}} & \textcolor[rgb]{ 1,  0,  0}{\textbf{0.258}} & \textcolor[rgb]{ 1,  0,  0}{\textbf{0.294}} & \cellcolor[HTML]{F6E9C5}\textcolor[rgb]{ 1,  0,  0}{\textbf{0.321}} & \cellcolor[HTML]{F6E9C5}\textcolor[rgb]{ 1,  0,  0}{\textbf{0.339}} \\
\cmidrule{2-19}          & \textbf{$\Delta$} & \textcolor{black}{{0.00\%}} & \textcolor{black}{{0.00\%}} & \textcolor{blue}{{0.21\%}} & \textcolor{blue}{{0.21\%}} & \textcolor{blue}{{1.23\%}} & \textcolor{blue}{{0.62\%}} & \textcolor{blue}{{2.04\%}} & \textcolor{blue}{{1.41\%}} & \textbf{$\Delta$} & \textcolor{black}{{0.00\%}} & \textcolor{black}{{0.00\%}} & \textcolor{black}{{0.00\%}} & \textcolor{black}{{0.00\%}} & \textcolor{black}{{0.00\%}} & \textcolor{blue}{{0.34\%}} & \textcolor{black}{{0.00\%}} & \textcolor{black}{{0.00\%}} \\
    \midrule
    \midrule
    \multirow{5}[0]{*}{\textbf{TimeBridge}} & {$96$} & \cellcolor[HTML]{F6E9C5}\textcolor[rgb]{ 1,  0,  0}{\textbf{0.414}} & \cellcolor[HTML]{F6E9C5}\textcolor[rgb]{ 1,  0,  0}{\textbf{0.444}} & \textcolor[rgb]{ 1,  0,  0}{\textbf{0.431}} & \textcolor[rgb]{ 1,  0,  0}{\textbf{0.456}} & \textcolor[rgb]{ 1,  0,  0}{\textbf{0.439}} & \textcolor[rgb]{ 1,  0,  0}{\textbf{0.462}} & \textcolor[rgb]{ 1,  0,  0}{\textbf{0.475}} & \textcolor[rgb]{ 1,  0,  0}{\textbf{0.492}} & {$96$} & \cellcolor[HTML]{F6E9C5}\textcolor[rgb]{ 1,  0,  0}{\textbf{0.149}} & \cellcolor[HTML]{F6E9C5}\textcolor[rgb]{ 1,  0,  0}{\textbf{0.203}} & 0.194 & 0.248 & 0.245 & 0.290 & 0.318 & 0.342 \\
          & {$192$} & 0.420 & 0.449 & \cellcolor[HTML]{F6E9C5}0.438 & \cellcolor[HTML]{F6E9C5}0.462 & 0.446 & 0.468 & 0.486 & 0.499 & {$192$} & 0.151 & 0.209 & \cellcolor[HTML]{F6E9C5}\textcolor[rgb]{ 1,  0,  0}{\textbf{0.192}} & \cellcolor[HTML]{F6E9C5}\textcolor[rgb]{ 1,  0,  0}{\textbf{0.247}} & \textcolor[rgb]{ 1,  0,  0}{\textbf{0.243}} & \textcolor[rgb]{ 1,  0,  0}{\textbf{0.287}} & \textcolor[rgb]{ 1,  0,  0}{\textbf{0.316}} & \textcolor[rgb]{ 1,  0,  0}{\textbf{0.338}} \\
          & {$336$} & 0.417 & 0.447 & 0.434 & 0.459 & \cellcolor[HTML]{F6E9C5}0.446 & \cellcolor[HTML]{F6E9C5}0.470 & \textcolor[rgb]{ 1,  0,  0}{\textbf{0.475}} & 0.495 & {$336$} & 0.157 & 0.217 & 0.198 & 0.254 & \cellcolor[HTML]{F6E9C5}0.247 & \cellcolor[HTML]{F6E9C5}0.292 & 0.321 & 0.342 \\
          & {$720$} & 0.428 & 0.454 & 0.443 & 0.465 & 0.456 & 0.475 & \cellcolor[HTML]{F6E9C5}0.511 & \cellcolor[HTML]{F6E9C5}0.514 & {$720$} & 0.167 & 0.227 & 0.208 & 0.262 & 0.258 & 0.298 & \cellcolor[HTML]{F6E9C5}0.329 & \cellcolor[HTML]{F6E9C5}0.347 \\
\cmidrule{2-19}          & \textbf{$\Delta$} & \textcolor{black}{{0.00\%}} & \textcolor{black}{{0.00\%}} & \textcolor{blue}{{1.60\%}} & \textcolor{blue}{{1.30\%}} & \textcolor{blue}{{1.57\%}} & \textcolor{blue}{{1.70\%}} & \textcolor{blue}{{7.05\%}} & \textcolor{blue}{{4.28\%}} & \textbf{$\Delta$} & \textcolor{black}{{0.00\%}} & \textcolor{black}{{0.00\%}} & \textcolor{black}{{0.00\%}} & \textcolor{black}{{0.00\%}} & \textcolor{blue}{{1.62\%}} & \textcolor{blue}{{1.71\%}} & \textcolor{blue}{{3.95\%}} & \textcolor{blue}{{2.59\%}} \\
    \midrule
    \midrule
    \multirow{5}[0]{*}{\textbf{TimeBase}} & {$96$} & \cellcolor[HTML]{F6E9C5}\textcolor[rgb]{ 1,  0,  0}{\textbf{0.627}} & \cellcolor[HTML]{F6E9C5}\textcolor[rgb]{ 1,  0,  0}{\textbf{0.551}} & \textcolor[rgb]{ 1,  0,  0}{\textbf{0.642}} & \textcolor[rgb]{ 1,  0,  0}{\textbf{0.559}} & \textcolor[rgb]{ 1,  0,  0}{\textbf{0.650}} & \textcolor[rgb]{ 1,  0,  0}{\textbf{0.567}} & 0.666 & 0.590 & {$96$} & \cellcolor[HTML]{F6E9C5}0.187 & \cellcolor[HTML]{F6E9C5}0.251 & 0.227 & 0.282 & 0.271 & 0.314 & 0.333 & 0.355 \\
          & {$192$} & 0.659 & 0.560 & \cellcolor[HTML]{F6E9C5}0.662 & \cellcolor[HTML]{F6E9C5}0.565 & 0.662 & 0.569 & \textcolor[rgb]{ 1,  0,  0}{\textbf{0.664}} & \textcolor[rgb]{ 1,  0,  0}{\textbf{0.586}} & {$192$} & 0.186 & \textcolor[rgb]{ 1,  0,  0}{\textbf{0.247}} & \cellcolor[HTML]{F6E9C5}\textcolor[rgb]{ 1,  0,  0}{\textbf{0.226}} & \cellcolor[HTML]{F6E9C5}\textcolor[rgb]{ 1,  0,  0}{\textbf{0.279}} & \textcolor[rgb]{ 1,  0,  0}{\textbf{0.269}} & \textcolor[rgb]{ 1,  0,  0}{\textbf{0.311}} & \textcolor[rgb]{ 1,  0,  0}{\textbf{0.331}} & \textcolor[rgb]{ 1,  0,  0}{\textbf{0.353}} \\
          & {$336$} & 0.686 & 0.576 & 0.686 & 0.577 & \cellcolor[HTML]{F6E9C5}0.694 & \cellcolor[HTML]{F6E9C5}0.585 & 0.702 & 0.606 & {$336$} & \textcolor[rgb]{ 1,  0,  0}{\textbf{0.185}} & 0.249 & \textcolor[rgb]{ 1,  0,  0}{\textbf{0.226}} & 0.281 & \cellcolor[HTML]{F6E9C5}0.270 & \cellcolor[HTML]{F6E9C5}0.312 & \textcolor[rgb]{ 1,  0,  0}{\textbf{0.331}} & 0.354 \\
          & {$720$} & 0.717 & 0.591 & 0.728 & 0.596 & 0.732 & 0.606 & \cellcolor[HTML]{F6E9C5}0.753 & \cellcolor[HTML]{F6E9C5}0.631 & {$720$} & 0.255 & 0.307 & 0.276 & 0.321 & 0.307 & 0.341 & \cellcolor[HTML]{F6E9C5}0.354 & \cellcolor[HTML]{F6E9C5}0.373 \\
\cmidrule{2-19}          & \textbf{$\Delta$} & \textcolor{black}{{0.00\%}} & \textcolor{black}{{0.00\%}} & \textcolor{blue}{{3.02\%}} & \textcolor{blue}{{1.06\%}} & \textcolor{blue}{{6.34\%}} & \textcolor{blue}{{3.08\%}} & \textcolor{blue}{{11.82\%}} & \textcolor{blue}{{7.13\%}} & \textbf{$\Delta$} & \textcolor{blue}{{1.07\%}} & \textcolor{blue}{{1.59\%}} & \textcolor{black}{{0.00\%}} & \textcolor{black}{{0.00\%}} & \textcolor{blue}{{0.37\%}} & \textcolor{blue}{{0.32\%}} & \textcolor{blue}{{6.50\%}} & \textcolor{blue}{{5.36\%}} \\
    \midrule
    \midrule
    \multirow{5}[0]{*}{\textbf{TimeEmb}} & {$96$} & \cellcolor[HTML]{F6E9C5}\textcolor[rgb]{ 1,  0,  0}{\textbf{0.387}} & \cellcolor[HTML]{F6E9C5}0.417 & \textcolor[rgb]{ 1,  0,  0}{\textbf{0.414}} & \textcolor[rgb]{ 1,  0,  0}{\textbf{0.433}} & \textcolor[rgb]{ 1,  0,  0}{\textbf{0.430}} & \textcolor[rgb]{ 1,  0,  0}{\textbf{0.442}} & \textcolor[rgb]{ 1,  0,  0}{\textbf{0.440}} & \textcolor[rgb]{ 1,  0,  0}{\textbf{0.461}} & {$96$} & \cellcolor[HTML]{F6E9C5}\textcolor[rgb]{ 1,  0,  0}{\textbf{0.166}} & \cellcolor[HTML]{F6E9C5}\textcolor[rgb]{ 1,  0,  0}{\textbf{0.222}} & \textcolor[rgb]{ 1,  0,  0}{\textbf{0.208}} & \textcolor[rgb]{ 1,  0,  0}{\textbf{0.258}} & \textcolor[rgb]{ 1,  0,  0}{\textbf{0.254}} & \textcolor[rgb]{ 1,  0,  0}{\textbf{0.292}} & 0.321 & \textcolor[rgb]{ 1,  0,  0}{\textbf{0.340}} \\
          & {$192$} & 0.388 & \textcolor[rgb]{ 1,  0,  0}{\textbf{0.415}} & \cellcolor[HTML]{F6E9C5}0.419 & \cellcolor[HTML]{F6E9C5}\textcolor[rgb]{ 1,  0,  0}{\textbf{0.433}} & 0.435 & \textcolor[rgb]{ 1,  0,  0}{\textbf{0.442}} & 0.454 & 0.466 & {$192$} & 0.168 & 0.224 & \cellcolor[HTML]{F6E9C5}0.210 & \cellcolor[HTML]{F6E9C5}0.260 & 0.255 & 0.294 & 0.320 & \textcolor[rgb]{ 1,  0,  0}{\textbf{0.340}} \\
          & {$336$} & 0.429 & 0.454 & 0.457 & 0.469 & \cellcolor[HTML]{F6E9C5}0.478 & \cellcolor[HTML]{F6E9C5}0.482 & 0.542 & 0.524 & {$336$} & 0.169 & 0.226 & 0.211 & 0.261 & \cellcolor[HTML]{F6E9C5}0.255 & \cellcolor[HTML]{F6E9C5}0.295 & 0.319 & \textcolor[rgb]{ 1,  0,  0}{\textbf{0.340}} \\
          & {$720$} & 0.401 & 0.429 & 0.425 & 0.442 & 0.444 & 0.453 & \cellcolor[HTML]{F6E9C5}0.470 & \cellcolor[HTML]{F6E9C5}0.484 & {$720$} & 0.171 & 0.230 & 0.212 & 0.264 & 0.256 & 0.296 & \cellcolor[HTML]{F6E9C5}\textcolor[rgb]{ 1,  0,  0}{\textbf{0.317}} & \cellcolor[HTML]{F6E9C5}\textcolor[rgb]{ 1,  0,  0}{\textbf{0.340}} \\
\cmidrule{2-19}          & \textbf{$\Delta$} & \textcolor{black}{{0.00\%}} & \textcolor{blue}{{0.48\%}} & \textcolor{blue}{{1.19\%}} & \textcolor{black}{{0.00\%}} & \textcolor{blue}{{10.04\%}} & \textcolor{blue}{{8.30\%}} & \textcolor{blue}{{6.38\%}} & \textcolor{blue}{{4.75\%}} & \textbf{$\Delta$} & \textcolor{black}{{0.00\%}} & \textcolor{black}{{0.00\%}} & \textcolor{blue}{{0.95\%}} & \textcolor{blue}{{0.77\%}} & \textcolor{blue}{{0.39\%}} & \textcolor{blue}{{1.02\%}} & \textcolor{black}{{0.00\%}} & \textcolor{black}{{0.00\%}} \\
    \bottomrule
    \end{tabular}}%
  \label{tab:main}%
\end{table*}%

\subsubsection{The Dominance of EF Paradigm (RQ1)}
In this section, we aim to demonstrate that optimal performance is not exclusively achieved by DF paradigm $L=H$ from the following results.

\textbf{Granular Numerical
Comparison.} For the perspective of numerical results, Table~\ref{tab:main} reveals a decisive advantage for the EF paradigm. For each specific $H$, the optimal results (highlighted in red) consistently less than the values on DF configurations (orange background, $L=H$), particularly at shorter output horizon spans ($L=96$ or $192$). The metric $\Delta$ indicates that EF yields relative precision improvements of up to 13.92\%. Notably, the synchronized DF setup constitutes only 23.20\% of the optimal cases in ETTh1 and 30.0\% in Weather. This evidence directly refutes the long-standing assumption that the output horizon must be synchronized with the evaluation horizon. 

\begin{figure*}[!t]
    \centering    \includegraphics[width=\linewidth]{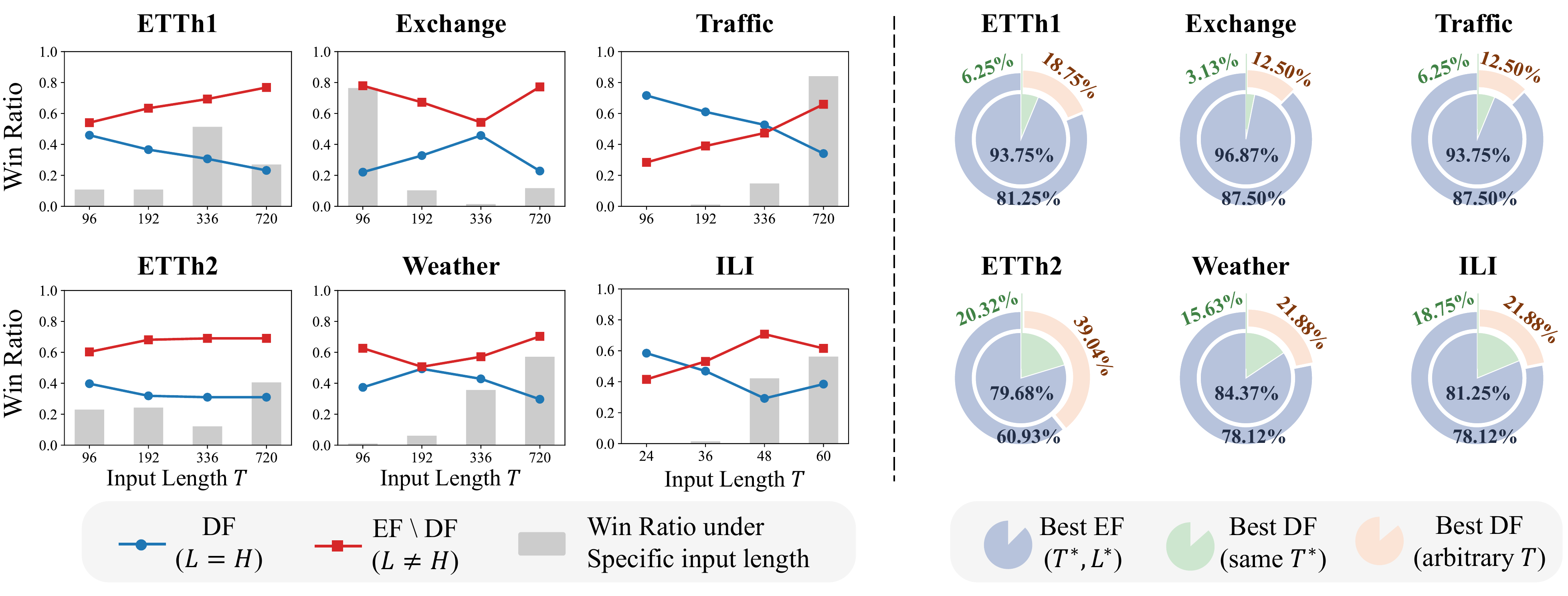}
    \caption{Statistical Dominance in the EF Paradigm. \textbf{LEFT}: We report the Win Ratio into the EF paradigm between the non-DF ($L \neq H$, red line) cases versus the conventional DF paradigm ($L=H$, blue line) across various input lengths $T$. The gray bars represent the marginal probability distribution of optimal performance relative to the input length $T$. \textbf{RIGHT}: We select the set of best input-output parameters $T^*,L*$ where each model perform as best as possible under the EF paradigm on all evaluation horizon. The win ratio is calculated between the DF with the same input length ($T^*$) and the DF with input length that perform best inside the DF paradigm.}
    \label{fig:placeholder}
\end{figure*}

\textbf{Global Statistical Comparison.} For the perspective of statistical results, Figure~\ref{fig:placeholder} (Left) further elucidates the scaling behaviour of this dominance. We observe that the Win Ratio of non-DF configurations (denoted as ``EF $\setminus$ DF", red line) remains dominant of all the input length $T$ at most times, frequently exceeding 80\%. While DF shows temporary leads at minimal $T$ (e.g., $T=96$ for Traffic and ILI), these regimes coincide with significantly low global performance, as indicated by the gray bars. On the one hand, these information-starved inputs fail to provide a sufficient receptive field for stable temporal reasoning in either DF or EF paradigm. On the other hand, the poor performance of EF is attributed to the excessive number of iterations required in the Pure-Extrapolation Phase in these low-$T$ settings, which compounds accumulated error. Conversely, in the Exchange dataset, where optimal performance concentrates at shorter $T$, EF still maintains a consistent lead, proving its structural robustness across diverse data distributions.

\subsubsection{``One-for-All" Generalization (RQ2)}

To evaluate the practical utility of our EF paradigm, we go beyond merely demonstrating marginal performance gains. Instead, we investigate a more profound question: \textbf{can a singular, well-calibrated EF model—configured with a fixed optimal pair $(T^*, L^*)$—robustly substitute an entire ensemble of specialized, re-customized, and retrained DF models?} Crucially, we seek to demonstrate that this superiority is not achieved by expanding the hyperparameter search space for every distinct output horizon. Rather, we aim to identify a universal ``One-for-All'' configuration that, once trained, achieves state-of-the-art performance across diverse horizons. This would fundamentally streamline the currently cumbersome LTSF benchmarking protocol, offering a minimalist yet powerful alternative to the prevailing re-train for every horizon' orthodoxy.

\textbf{Granular Numerical
Comparison.} As shown in Table~\ref{tab:main}, we observe that on both the ETTh1 and Weather datasets, when the input length is 720, the EF paradigm nearly always identifies a single optimal input length $T^*$ that requires only once training, corresponding to the neatly aligned red-highlighted performance row in the table, that outperforms the dedicated DF paradigm ($L=H$) across all evaluation horizons $H$, achieving equivalent or superior performance in 94.5\% of the metrics listed in the table.

\textbf{Global Statistical Comparison.} To demonstrate that the above conclusions are not merely selected examples, we aim to show that this phenomenon is reflected across all datasets and all baselines. Hence we further quantify it in Figure~\ref{fig:placeholder} (Right), where we define ``Best EF" as the configuration with the highest cumulative of optimal evaluation metrics (using smaller $T, L$ as tie-breakers) and denoted the input length and output horizon of ``Best EF" are $T^*,L^*\in\{96, 192, 336, 720\}$ (ILL with $\{24,36,48,60\}$). We compared it with two DF paradigms by Win Ratio\footnote{Since the EF paradigm does not require retraining, it is better than the DF paradigm under identical metric results.}: \ding{202} \textbf{Same Input Length (Inner Ring):} Compared against DF models trained with identical input lengths ($T=T^*$), the ``Best EF" model maintains an absolute lead, exceeding an 80\% win rate across all datasets and even reaching 96.87\% on Exchange. This confirms that with identical data access, evolutionary reasoning of EF is intrinsically superior to monolithic projection of DF. \ding{203} \textbf{Optimal Input Length (Outer Circle):} When challenged by the "best-case" DF models selected from arbitrary optional $T$ for each evaluation horizon $H$, our singular EF operator remains victorious in all the cases, peaking at an 87.50\% superiority rate. 

These results provide a striking conclusion: a singular model with EF sufficient generalization capacity to navigate common future horizons with higher fidelity than a collection of task-aligned, specialized and retrained DF models. This not only validates our Decoupling Hypothesis but also offers a pivotal paradigm shift for large-scale LTSF deployment, reducing computational overhead by replacing multi-model ensembling evaluation. 

\begin{figure*}[!t]
    \centering
    \includegraphics[width=\linewidth]{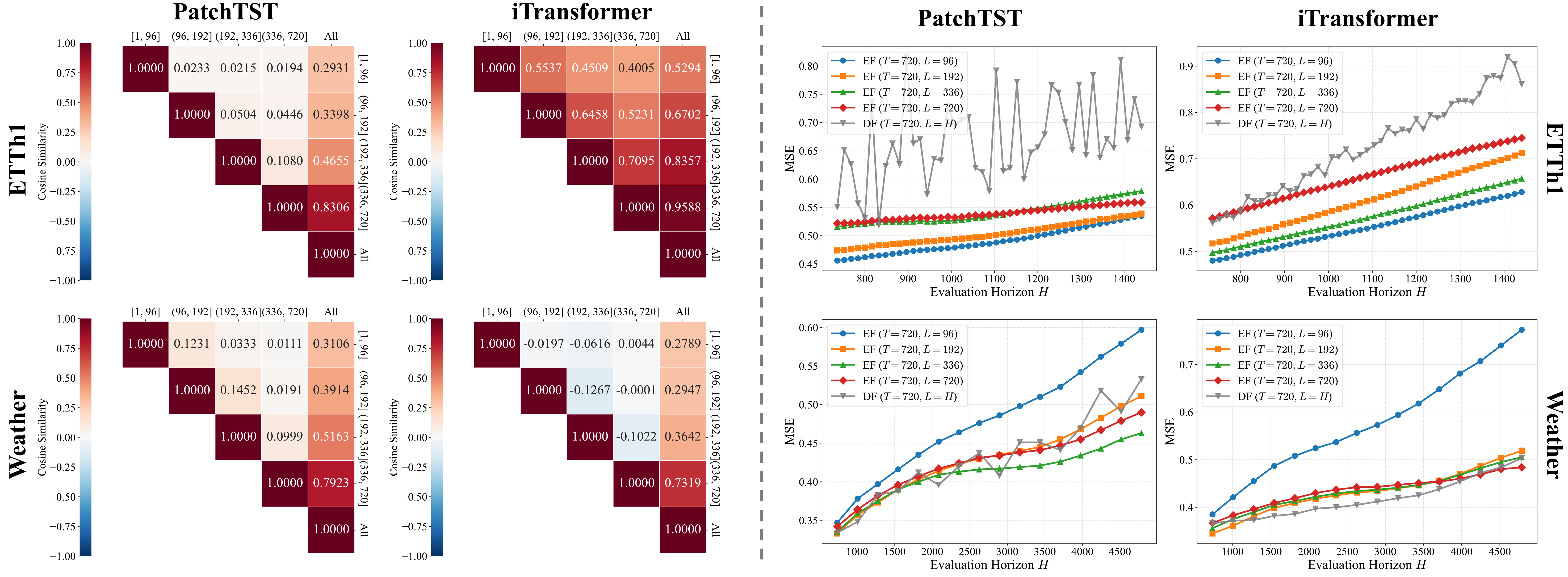}
    \caption{\textbf{LEFT}: Visualization of Gradient Conflict and Distal Dominance in DF paradigm on $T=L=720$. We compute the average cosine similarity of gradients across partitioned forecasting segments ($[1, 96]$ to $(336, 720]$) and the total sequence (``All") during training. \textbf{RIGHT}: Extreme LTSF on $T=720$.}
    \label{fig:gradient}
\end{figure*}

\subsection{Extreme Extrapolation Stability (RQ3)}\label{sec:ees}

Having unveiled the latent extrapolation capabilities of contemporary models originally designed for DF under our EF paradigm, we are naturally compelled to probe the upper boundaries of this potential: \textbf{What is the asymptotic limit of the EF paradigm?} In other words, how does EF perform under extremely long-range extrapolation scenarios? Does the performance advantage of the EF paradigm over the DF paradigm persist? 

To conduct this stress test, we evaluate PatchTST and iTransformer on the ETTh1 and Weather datasets. We fix the input length $T=720$ and select all the output horizons $L \in \{96, 192, 336, 720\}$. We then stretch the evaluation horizon $H$ beyond standard benchmarks, pushing $H$ to the asymptotic limit where the validation set contains only a single valid sample (i.e., $T+H$ equals the total available time steps in validation set). As illustrated in Figure~\ref{fig:gradient} (Right), the performance degradation of EF remains remarkably smooth and controlled. In ETTh1, all EF configurations overwhelmingly outperform the task-aligned DF paradigm at extreme horizons, with the performance gap monotonically widening as $H$ increases. In Weather, while the highly-localized input ($L=96$) eventually decays due to insufficient periodicity coverage, configurations with slightly larger output horizons maintain robust stability, consistently outperforming horizon-specific retrained DF models. This empirical evidence culminates in a profound conclusion: the error accumulation inherent in the EF reasoning process is significantly less detrimental than dilemmas crippling the DF paradigm at extreme scales.

\subsection{Optimization Pathology: Gradient Conflict (RQ4)}\label{sec:opt_path}

To fundamentally elucidate why the EF paradigm achieves superior performance at drastically shorter output horizons, we shift our lens from representational capacity to the underlying optimization dynamics. We posit that long output horizon DF paradigm inherently suffers from a optimization pathology, where the model is compelled to reconcile distinct dynamical patterns across near-term, mid-term, and distant horizons within a shared parameter space.

To validate this, we visualize the cosine similarity matrices between the gradient directions of partitioned temporal segments $[1,96]$, $(97,192]$, $(192,336]$, $(336, 720]$ and the truth optimization trajectory (denoted as ``All") during training in Figure~\ref{fig:gradient}. Our empirical analysis reveals two striking phenomena that paralyze the conventional DF paradigm: \ding{202} \textbf{Adversarial Gradient Conflict:} We observe that the gradient directions of different forecasting segments are not merely distinct but often conflicting. The similarity scores between different segments are predominantly near-orthogonal, with certain segment pairs even exhibiting negative cosine similarity. This indicates that the parameter updates required to capture deterministic near-term patterns are geometrically incompatible or even adversarial. \ding{203} \textbf{Monotonic Distal Hijacking:} Simultaneously, we observe that the cosine similarity between individual segment gradients and the total gradient ``All'' exhibits a strictly monotonic increase with respect to the temporal distance. Specifically, the near-term segment shows the lowest alignment with the final gradient direction, while the distal-term segment shows the highest. Corroborating the performance collapse shown in Figure~\ref{fig:first_truncated}, this confirms that the optimization trajectory is effectively hijacked by the distal gradients, leading to a chronic under-fitting of near-term time steps. These findings provide a compelling justification for the EF paradigm: by strictly decoupling the output horizon, EF acts as an optimization filter, isolating local dynamics from the interference of distal gradients. Additional case studies and further analytical exploration on this topic are in Appendix~\ref{app:optpath}.

\section{Conclusion}
This work challenges the monolithic orthodoxy of current LTSF research by introducing the Evolutionary Forecasting (EF) paradigm. We unmask the prevailing Direct Forecasting (DF) paradigm is not merely an optimization trap characterized by severe gradient conflicts. Specifically, we fundamentally dismantle the rigid convention coupling output and evaluation horizons, thereby unlocking the latent extrapolation capabilities that have long remained dormant in modern LTSF architectures. Our contribution extends beyond offering new parametric freedom; we pioneer a ``One-for-All'' generalization, demonstrating that a singular EF model exhibits robust asymptotic stability, consistently outperforming corresponding ensembles of re-customized and re-trained DF baselines across both common and extreme evaluation horizon. Ultimately, this work propels the LTSF field beyond the limitations of passive ``Static Mapping" steering it toward autonomous ``Evolutionary Reasoning''. We sincerely hope that our comprehensive studies will catalyze not only new insights into model design but also foster the establishment of a more efficient protocol for benchmarking within the community.


\section*{Impact Statement}

This paper presents work whose goal is to advance the field of Machine Learning. Time series forecasting serves as a cornerstone in critical societal domains, including energy grid management, weather forecasting, and economic planning. By significantly enhancing the performance of deep learning-based forecasting models, our proposed Evolutionary Forecasting paradigm facilitates more reliable deployments in these real-world applications. We do not foresee any specific negative societal consequences of our work.


\nocite{langley00}

\bibliography{example_paper}
\bibliographystyle{icml2026}

\newpage
\appendix
\onecolumn

   
   

\section{Pseudo-code of Evolutionary Forecasting}\label{app:pseudocode}
We present the complete algorithmic workflow of the proposed Evolutionary Forecasting (EF) paradigm in Algorithm~\ref{alg:ef_framework}. 

\begin{algorithm}[!h]
   \caption{Evolutionary Forecasting (EF) Training \& Inference}
   \label{alg:ef_framework}
\begin{algorithmic}
   \STATE {\bfseries Require:} Historical observation  $T$, Model output horizon $L$.
   \STATE {\bfseries Model:} Historical observation $\mathbf{X} \in \mathbb{R}^{T \times C}$, Evaluation Horizon $H$.
   
   \STATE \rule{\linewidth}{0.5pt}
   \STATE \textbf{Stage I: Optimization (Training)}
   \STATE \textit{\textcolor[HTML]{214D7B}{// Objective: Minimize error strictly on the Direct Phase ($k=1$), equivalent to Teacher Forcing.}}
   \STATE {\bfseries Input:} Historical observation $\mathbf{X} \in \mathbb{R}^{T \times C}$
   \WHILE{not converged}
       \STATE Sample batch $(\mathbf{X}, \mathbf{Y})$ where $\mathbf{Y} \in \mathbb{R}^{L \times C}$.
       \STATE $\hat{\mathbf{Y}} = \mathcal{F}_\theta(\mathbf{X})$ \COMMENT{\textcolor[HTML]{098842}{\# Direct Forecasting}}
       \STATE $\mathcal{L} = \text{MSE}(\hat{\mathbf{Y}}, \mathbf{Y})$
       \STATE Update $\theta \leftarrow \theta - \eta \nabla_\theta \mathcal{L}$
   \ENDWHILE
   \STATE \textbf{Output:} Optimized parameters $\theta^*$.

   \STATE \rule{\linewidth}{0.5pt}
   \STATE \textbf{Stage II: Evolutionary Inference (Testing)}
   \STATE \textit{\textcolor[HTML]{214D7B}{// Objective: Generate arbitrary horizon $H$ via Block-wise Reasoning.}}
   \STATE {\bfseries Input:} Test observation $\mathbf{X} \in \mathbb{R}^{T \times C}$, Evaluation Horizon $H$.
   \STATE Calculate reasoning steps: $K = \lceil H / L \rceil$
   \STATE Initialize cumulative prediction: $\hat{\mathcal{Y}}_{accum} = [\;]$
   
   \FOR{$k = 1$ {\bfseries to} $K$}
       \STATE \textit{\textcolor[HTML]{214D7B}{// 1. Construct Input $\tilde{\mathbf{X}}^{(k)}$ (Definition \ref{def:barltsf})}}
       \IF{$k = 1$}
           \STATE \textbf{\textcolor{red}{\# Direct Phase:}} 
           \STATE $\tilde{\mathbf{X}}^{(k)} = \mathbf{X}$
       \ELSIF{$1 < k < T/L + 1$}
           \STATE \textbf{\textcolor{red}{\# Semi-Extrapolation Phase:}}
           \STATE $\mathbf{X}_{part} = \mathbf{X}_{(k-1)L : T}$ 
           \STATE $\hat{\mathbf{Y}}_{part} = \text{Concat}(\hat{\mathcal{Y}}_{accum})_{1:(k-1)L}$
           \STATE $\tilde{\mathbf{X}}^{(k)} = \text{Concat}(\mathbf{X}_{part}, \hat{\mathbf{Y}}_{part})$
       \ELSE
           \STATE \textbf{\textcolor{red}{\# Pure-Extrapolation Phase:}}
           \STATE $\tilde{\mathbf{X}}^{(k)} = \text{Concat}(\hat{\mathcal{Y}}_{accum})_{(k-1)L - T : (k-1)L}$
       \ENDIF
       
       \STATE \textit{\textcolor[HTML]{214D7B}{// 2. Forecasting }}
       \STATE $\mathbf{B}^{(k)} = \mathcal{F}_{\theta^*}(\tilde{\mathbf{X}}^{(k)})$
       \STATE Append $\mathbf{B}^{(k)}$ to $\hat{\mathcal{Y}}_{accum}$
   \ENDFOR
   
   \STATE Finalize: $\hat{\mathbf{Y}} = \text{Concat}(\hat{\mathcal{Y}}_{accum})_{1:H}$
   \STATE {\bfseries Return} $\hat{\mathbf{Y}}$
\end{algorithmic}
\end{algorithm}

\section{Related Work}\label{app:relaterwork}

\subsection{Paradigms in Multi-step Forecasting: From Recursion to Direct Forecasting}
The challenge of multi-step ahead forecasting has been rigorously studied for decades, oscillating between two primary paradigms: \textit{Recursive} (iterative) and \textit{Direct} strategies~\cite{taieb2012review, sorjamaa2007methodology}. The representative of Recursive strategies is Auto Regression, which trains a one-step model, like ARIMA~\cite{box1976analysis,box2015time} or RNNs~\cite{salinas2020deepar}, and use predicted values as inputs for subsequent steps, theoretically allowing infinite horizons but suffering from error accumulation~\cite{lim2021time,torres2021deep}, Direct strategies train specific models for specific horizons to minimize bias but entail higher computational costs and imply independence between steps~\cite{chevillon2007direct}.  Our proposed Evolutionary Forecasting (EF) revisits the Recursive strategy through a modern lens. Unlike the classic point-wise recursion, EF adopts a Block-wise Recursive mechanism ($L > 1$). This hybrid approach balances the stability of direct mapping (within the block) with the extrapolation capability of recursion (across blocks), effectively acting as a strategic choice within the horizon-dependent taxonomy emphasized by TimeRecipe~\cite{zhao2025timerecipe}.

\subsection{The Rigidity of Direct Forecasting and Architectural Adaptations}
The instability of recursive RNNs established the dominance of the Direct Forecasting (DF) paradigm~\cite{zhou2021informer}, refined by efficient architectures like DLinear~\cite{zeng2023transformers} and PatchTST~\cite{nie2023time}. However, DF imposes a rigid coupling between output and evaluation horizons. To alleviate this rigidity and instability, recent works have resorted to architectural modifications or auxiliary regularizations.
For instance, D2Vformer~\cite{song2025d2vformer} introduces a flexible Time-Position Embedding within a decoder-like structure to enable variable-length generation;
KOSS~\cite{wang2025koss} utilizes Selective State Space Models to stabilize dependencies;
and PRISM~\cite{chen2025prism} constrains autoregressive errors via regularization.
Distinct from these approaches, EF operates at a higher level of abstraction. Rather than designing ad-hoc architectural components, we unmask and unlock the latent extrapolation capabilities that are already inherent in standard DF backbones but suppressed by the rigid $L=H$ convention. We identify that the bottleneck is not architectural capacity, but an optimization pathology—specifically, the severe underfitting caused by gradient conflicts in long-horizon training. By resolving this optimization dilemma, EF achieves superior asymptotic stability and flexibility without altering the underlying model architecture.

\subsection{Universal Enhancement Methods in LTSF}
Beyond specific model architectures, researchers have sought universal strategies to enhance forecasting performance. To distinguish EF's positioning, we categorize these into optimization-centric and module-based approaches. \ding{202} \textbf{Optimization-objective Enhancements.} A parallel line of research focuses on refining the loss landscape to guide model convergence better. Comprehensive surveys~\cite{jadon2024comprehensive} emphasize the utility of robust objectives like Huber Loss~\cite{qiu2025duet} for handling outliers. Recent innovations move beyond point-wise error. For examples, FreDF~\cite{wang2024fredf} shifts optimization to the frequency domain to capture spectral dependencies, DBLoss~\cite{qiu2025dbloss} incorporates decomposition priors to align trend and seasonality learning, and Patch-wise Structural Loss~\cite{kudrat2025patch} explicitly penalizes morphological distortions. These methods aim to improve performance by designing more sophisticated supervision targets. \ding{203} \textbf{Structural Module Enhancements.} Additionally, researchers have introduced auxiliary structural modules that can be plugged into various backbones to tackle the perennial challenge of distribution shift~\cite{liu2022non}. For examples, RevIN~\cite{kim2021reversible} and Dish-TS~\cite{fan2023dish} propose reversible instance normalization layers as universal paradigms to align training and inference distributions, serving as effective patches for distribution drift. However, our EF fundamentally diverges from these additive approaches. First, unlike optimization-centric methods that introduce complex auxiliary loss functions, EF retains the standard objective but resolves the near-term under-fitting dilemma. We identify that the bottleneck is not the loss function itself, but the gradient conflicts inherent in long-horizon optimization (Section~\ref{sec:opt_path}). Second, unlike structural methods that append extra parameters or modules (e.g., Normalization layers), EF requires no architectural modifications.Instead of ``adding'' external capacity, EF excavates the inherent, ignored extrapolation capabilities already present in baseline models. We demonstrate that by simply correcting the optimization dynamics (via short-horizon training), models can effectively handle distribution shifts and achieve SOTA performance without auxiliary prosthetics.

\subsection{Generative Foundation Models in LTSF} The resurgence of autoregressive principles is most prominently validated by the emerging wave of Foundation Models in LTSF, where the generative paradigm has effectively become the mainstream~\cite{jin2023time,cao2023tempo,liu2024timer,wang2024timemixer++,ansari2024chronos,das2024decoder}. One stream of research leverages the pre-existing reasoning capabilities of Large Language Models (LLMs), such as Time-LLM~\cite{jin2023time} and Tempo~\cite{cao2023tempo}, which align time series data with textual modalities. Another stream constructs native time series foundation models, exemplified by Timer-XL~\cite{liu2024timer} and TimeMixer++~\cite{wang2024timemixer++}, designed for universal pattern mining. Notably, TimesFM~\cite{das2024decoder} explicitly employs a decoder-only architecture with patch-wise (block-wise) inference, echoing the chunked reasoning spirit of our EF paradigm. However, our work distinguishes itself in both scope and optimization philosophy. First, distinct from the pursuit of universality via massive pre-training, we focus on unlocking the latent extrapolation potential of normal deep learning architectures (e.g., iTransformer~\cite{liu2023itransformer}, PatchTST~\cite{nie2023time}) that are traditionally constrained by the rigid DF paradigm. Second, and most crucially, we contribute a fundamental optimization insight: we demonstrate that these models do not require training on long horizons to achieve robust long-term capabilities. By proving that short-horizon training coupled with EF suffices for asymptotic stability, we offer a cost-effective alternative to the ``long-train-for-long-predict'' dogma, providing profound implications for future model design and training efficiency in the broader LTSF community.

\subsection{Block-wise Generative Reasoning across Modalities}
The philosophy of generating sequences in ``chunks'' rather than single units parallels recent advancements in broader generative domains. In Computer Vision, models like MaskGIT~\cite{chang2022maskgit} generate images via iterative patch in-painting; in NLP, techniques like Block-wise Parallel Decoding~\cite{stern2018blockwise} and Medusa~\cite{cai2024medusa} accelerate inference by predicting multiple tokens simultaneously; similarly, Block Diffusion~\cite{arriola2025block} interpolates between autoregressive and diffusion models. Our EF paradigm aligns with this broader generative evolution. By adapting this Block-wise Reasoning intuition to LTSF, EF successfully unmasks and unlocks the latent extrapolation capabilities inherent in modern time series models—capabilities that were previously suppressed by the rigid, monolithic nature of the traditional Direct Forecasting orthodoxy. This suggests that the effectiveness of intermediate-scale reasoning is a universal principle shared across modalities.

\subsection{Clarification on ``Evolutionary" Terminology}
We clarify that our term ``Evolutionary" in EF refers to the temporal evolution of dynamical systems—specifically, the progressive unfolding of future states via iterative reasoning block. This is distinct from Evolutionary Algorithms~\cite{bartz2014evolutionary} or Genetic Algorithms~\cite{holland1992genetic} used for hyperparameter optimization or architecture search in forecasting literature, such as ES-LSTM~\cite{yingnan2025hybrid} or ELATE~\cite{murray2025elate}. EF focuses on the evolution of the inference process, not the optimization of the model population.

\section{Justification}\label{app:justification}

To clarify the theoretical advantages of the proposed Evolutionary Forecasting (EF) paradigm, we provide a rigorous mathematical comparison with the traditional Point-wise Autoregressive (AR) paradigm. Traditional Deep Autoregressive models~\cite{salinas2020deepar} typically operate under a Point-wise Recursive mechanism. We formally define this paradigm using a parallel notation system to highlight the structural differences. Let $\mathcal{A}_\phi: (\mathbb{R}^{1 \times C}, \mathbb{R}^{d_h}) \to (\mathbb{R}^{1 \times C}, \mathbb{R}^{d_h})$ be a stateful autoregressive model (e.g., an RNN cell), where $d_h$ denotes the dimension of the hidden state. Given the same evaluation horizon $H$, the AR paradigm generates the prediction $\hat{\mathbf{Y}} \in \mathbb{R}^{H \times C}$ sequence-point by sequence-point. The prediction at step $t$ ($1 \le t \le H$), denoted as $\hat{\mathbf{y}}_t \in \mathbb{R}^{1 \times C}$, is formulated as:
\begin{equation}
    \hat{\mathbf{y}}_t, \mathbf{h}_t = \mathcal{A}_\phi(\hat{\mathbf{y}}_{t-1}, \mathbf{h}_{t-1}),
\end{equation}
where $\hat{\mathbf{y}}_{t-1}$ is the output from the previous step (with $\hat{\mathbf{y}}_0 = \mathbf{x}_T$), and $\mathbf{h}_{t-1}$ is the hidden state carrier passed from the previous step. The fundamental distinction lies in how the two paradigms manage \textbf{context}. We distinguish EF from AR as a transition from Implicit State Recursion to Explicit Window Evolution.

\textbf{AR (Stationary View).} The AR model $\mathcal{A}_\phi$ is spatially ``stationary.'' It does not re-read the historical sequence $\mathbf{X}$. Instead, it relies on the hidden state $\mathbf{h}_t$ to act as a compressed memory of the entire history.
\begin{equation}
    \text{Context}_{\text{AR}}(t) = \mathbf{h}_{t-1} \quad (\text{Implicit, Compressed}).
\end{equation}
This compression often leads to ``information forgetting'' in long-horizon forecasting, as the state $\mathbf{h}_t$ must encode an increasingly long history into a fixed-size vector.

\textbf{EF (Sliding View).} The EF model $\mathcal{F}_\theta$ is ``dynamic.'' At every reasoning block $k$, EF \textbf{explicitly} reconstructs a full physical window $\tilde{\mathbf{X}}^{(k)}$.
\begin{equation}
    \text{Context}_{\text{EF}}(k) = \tilde{\mathbf{X}}^{(k)} \in \mathbb{R}^{T \times C} \quad (\text{Explicit, Lossless Window}).
\end{equation}
By sliding the window forward (as defined in the $k$-Block definition), EF ensures that the model always has access to the most recent $T$ distinct raw features, avoiding the bottleneck of hidden state compression.

    


\section{Experiments}\label{app:exp}

\textbf{Training Protocol.} To strictly isolate the performance gains attributed to the decoupling of $L$ and evaluation $H$, we intentionally exclude complex optimization strategies such as scheduled sampling~\cite{bengio2015scheduled} or curriculum learning~\cite{bengio2009curriculum}. Instead, we uniformly employ the standard Teacher Forcing strategy. Specifically, the model in EF is optimized exclusively on the $1$-st Reasoning Block (the Direct Phase, $k=1$). This protocol treats the short-horizon forecast as a standard supervised learning task, mathematically equivalent to applying DF on a reduced window $L$ during training. This ensures that any observed improvements stem from the inherent superiority of the EF paradigm rather than advanced training tricks. 

\subsection{Main Experiments}\label{app:mainexp}

In this section, we provide a comprehensive report of the experimental results across six real-world datasets. The detailed performance metrics (MSE and MAE) are exhaustively listed in Table~\ref{tab:dlinear_all} (DLinear), Table~\ref{tab:patchtst_all} (PatchTST), Table~\ref{tab:itransformer_all} (iTransformer), Table~\ref{tab:timemier_all} (TimeMixer), Table~\ref{tab:sparsetsf_all} (SparseTSF), Table~\ref{tab:timebridge_all} (TimeBridge), Table~\ref{tab:timebase_all} (TimeBase) and Table~\ref{tab:timeemb_all} (TimeEmb). These tables cover all model configurations with multiple input length $T$, output horizon $L$ and evaluation horizon $H$ sweeping through $\{96, 192, 336, 720\}$ (for ILL, $T,L,H \in \{24, 36, 48, 60\}$). To facilitate a granular analysis of the performance landscape, we employ the following visual formatting in our tables: \ding{202} \textbf{Direct Forecasting (DF) Baseline:} Results obtained under the conventional DF paradigm, where the output horizon is rigidly coupled to the evaluation horizon ($L=H$), are highlighted with an \colorbox[HTML]{F6E9C5}{Orange Background}. \ding{203} \textbf{Local Best:} For a specific input length $T$ and evaluation horizon $H$, the best performing metrics are marked with \underline{\textcolor{blue}{Underline}}. \ding{204} \textbf{Global Optimum:} Across all tested input lengths $T$ for a fixed evaluation horizon $H$, the globally optimal performance is denoted in \textbf{\textcolor{red}{Bold}}. This indicator helps identify the absolute best configuration for a given task. A thorough examination of the experimental results reveals three critical insights regarding the behavior of EF versus the conventional DF paradigm:

\textbf{The Dominance of Decoupled Reasoning.} 
It is immediately observable that configurations utilizing the proposed EF paradigm ($L \neq H$) consistently outperform the rigid DF baseline ($L=H$). In the vast majority of cases, both the local best performances (\underline{\textcolor{blue}{Underlined}}) and the global optima (\textbf{\textcolor{red}{Bold}}) fall outside the orange-shaded DF rows. This empirical evidence validates the fundamental superiority of Evolutionary Forecasting, demonstrating that decoupling the training horizon from the evaluation target is a more effective strategy for long-term time series modeling.

\textbf{Input Length and Extrapolation Stability.} 
With the exception of the Exchange dataset, we observe a distinct divergence trend: the performance gap between EF and DF widens significantly as the input length $T$ increases. This phenomenon suggests that EF effectively leverages extended historical context to stabilize its block-wise extrapolation. In contrast, the DF paradigm struggles to utilize longer inputs effectively, likely due to the optimization difficulties inherent in mapping long-history inputs directly to long-horizon outputs without intermediate supervision.

\textbf{The Trade-off in Limited Context Regimes.} 
We note that in specific short-input scenarios (e.g., small $T$ on ETTh1), the conventional DF paradigm ($L=H$) occasionally achieves competitive performance. We attribute this anomaly to the confluence of contextual insufficiency and recursion risks. Specifically, when $T$ is short, the scarcity of historical information restricts any model from learning complex dynamics, placing the system in a high-bias state where Global Optima (\textbf{\textcolor{red}{Red Bold}}) are rarely achieved. Simultaneously, for EF, a short input combined with a long horizon necessitates a high number of reasoning steps ($K = \lceil H/L \rceil$) dominated by the Pure-Extrapolation Phase, thereby increasing the risk of error accumulation. However, as the input length $T$ increases, this trade-off resolves decisively in favor of EF. The increased context allows EF to better anchor its recursive blocks, avoiding the optimization pathology that plagues DF. Consequently, the highest quality predictions are almost exclusively achieved by EF models with sufficient input context, underscoring the necessity of our ``To See Far, Look Close'' strategy supported by adequate historical observation.

\subsection{Optimization Pathology of Direct Forecasting}\label{app:optpath}

To rigorously diagnose the root cause of the performance degradation observed in DF paradigm under long-horizon settings, we investigate the optimization dynamics from a gradient perspective. We hypothesize and reveal a critical phenomenon: Gradient Conflict and Distal Hijacking. Specifically, optimization objectives from different future time steps pull the model parameters in conflicting directions, and the magnitude of gradients is often unevenly distributed, leading to optimization instability. We c 
conduct these experimental exploration on representative ETTh1 and Weather datasets of all baselines models under the configuration of $T=L=720$.

\subsubsection{Multi-Scale Gradient Analysis Framework} 
Concretely, we propose a Multi-Scale Gradient Analyzer to dissect the training dynamics. Let $\mathcal{D} = \{(\mathbf{X}_b, \mathbf{Y}_b)\}_{b=1}^B$ denote the dataset consisting of $B$ batches per epoch. For a DF model $\mathcal{F}_\theta$ parameterized by $\theta$, the prediction for an input $\mathbf{X}$ is $\hat{\mathbf{Y}} = \mathcal{F}_\theta(\mathbf{X}) \in \mathbb{R}^{H \times C}$. We partition the total output horizon $L$ (e.g., $L=720$) into a set of disjoint intervals (segments) plus the complete horizon. Based on our experimental setup, we define the segments as:
\begin{equation}
    \mathcal{S} = \Big\{ \underbrace{[1, 96]}_{s_{96}}, \underbrace{(96, 192]}_{s_{192}}, \underbrace{(192, 336]}_{s_{336}}, \underbrace{(336, 720]}_{s_{720}}, \underbrace{[1, 720]}_{s_{\text{all}}} \Big\}.
\end{equation}
For a specific batch $b$ in epoch $e$, and a specific segment $s \in \mathcal{S}$, we denote the segment-specific loss as $\mathcal{L}_s(\theta; \mathbf{X}_b, \mathbf{Y}_b)$. The gradient vector corresponding to this segment is computed via back-propagation:
\begin{equation}
    \mathbf{g}_{s}^{(b,e)} = \nabla_\theta \mathcal{L}_s(\theta; \mathbf{X}_b, \mathbf{Y}_b) \in \mathbb{R}^{|\theta|},
\end{equation}
where $\mathbf{g}_{s}^{(b,e)}$ is the flattened vector of all learnable parameters $\theta$ derived solely from the supervision signal of segment $s$. Note that $\mathbf{g}_{\text{all}}^{(b,e)}$ represents the actual update direction applied to the model in standard DF training.

\subsubsection{Evaluation Metrics for Optimization Pathology}

To quantify the conflict and dominance between different temporal segments, we introduce two key metrics:

\textbf{Metric I: Gradient Direction Alignment (Cosine Similarity).}
This metric measures the geometric consistency between the optimization directions of two segments $i, j \in \mathcal{S}$. For a batch $b$ in epoch $e$, the pairwise cosine similarity is defined as:
\begin{equation}
    \text{Sim}_{i,j}^{(b,e)} = \text{Cos}(\mathbf{g}_i^{(b,e)}, \mathbf{g}_j^{(b,e)}) = \frac{\mathbf{g}_i^{(b,e)} \cdot \mathbf{g}_j^{(b,e)}}{\|\mathbf{g}_i^{(b,e)}\|_2 \|\mathbf{g}_j^{(b,e)}\|_2},
\end{equation}
where $\|\cdot\|_2$ denotes the $L_2$ norm. A value close to $1$ indicates alignment, while a value close to $0$ (orthogonality) or $-1$ (opposition) indicates severe \textbf{Gradient Conflict}, suggesting that optimizing for segment $i$ may degrade the performance on segment $j$ for all $i,j=1,2,\dots,S$.

\textbf{Metric II: Gradient Norm Contribution (Norm Ratio).}
This metric quantifies the contribution of a specific segment $s$ to the total parameter update. The norm ratio for segment $s$ relative to the global update is defined as:
\begin{equation}
    R_s^{(b,e)} = \frac{\|\mathbf{g}_s^{(b,e)}\|_2}{\|\mathbf{g}_{\text{all}}^{(b,e)}\|_2}.
\end{equation}
An imbalanced $R_s$ distribution (e.g., $R_{720} \gg R_{96}$) indicates \textbf{Distal Dominance}, where the optimization is hijacked by distal errors, causing the model to underfit local dynamics.

\subsubsection{Statistical Aggregation and Visualization}
\begin{figure*}[!t]
    \centering
    \includegraphics[width=\linewidth]{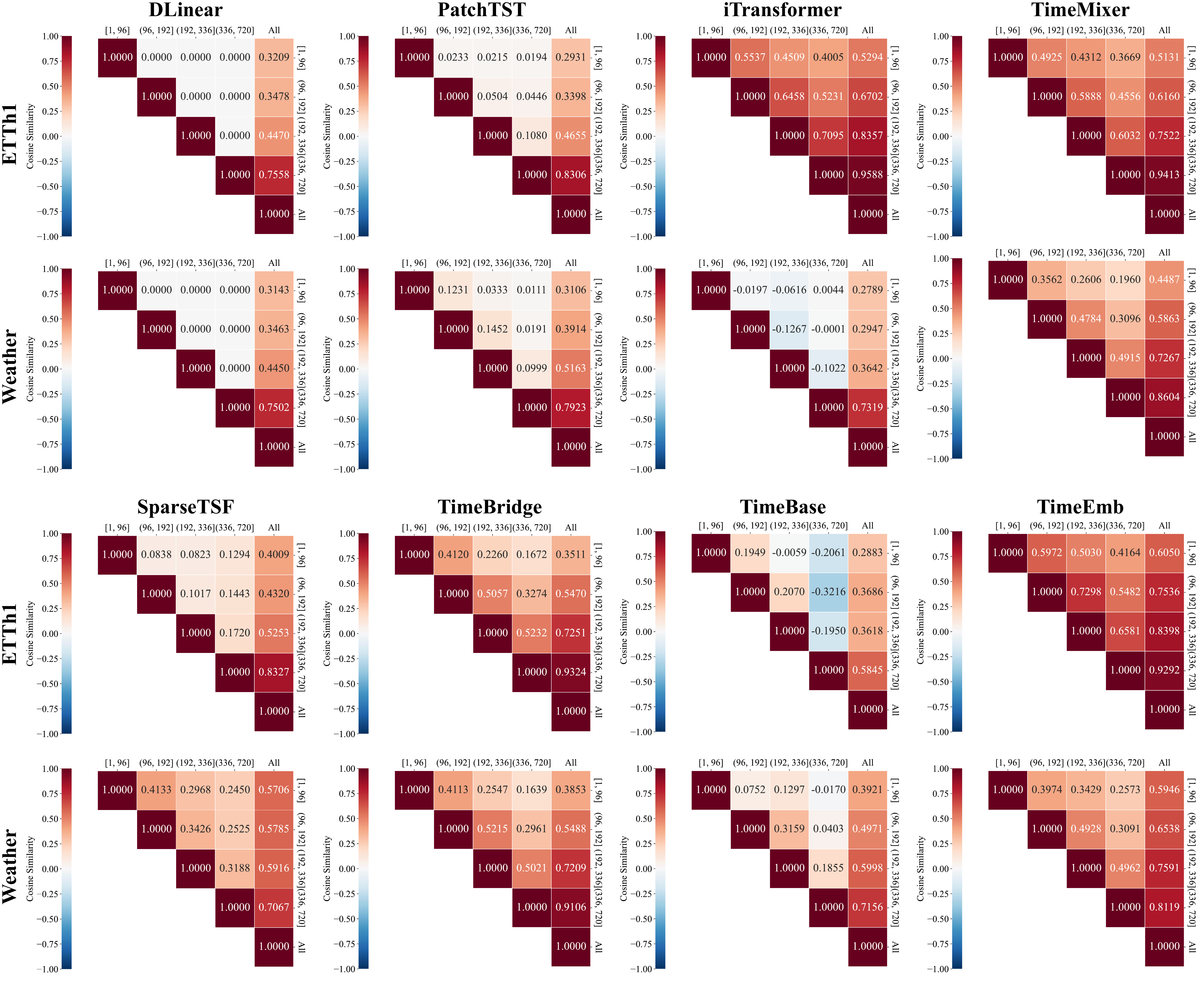}
    \caption{Visualization of Gradient Conflict and Distal Dominance in DF paradigm on $T=L=720$. We compute the average cosine similarity of gradients across partitioned forecasting segments ($[1, 96]$ to $(336, 720]$) and the total sequence (``All") during the training of all baselines on ETTh1 and Weather.}
    \label{fig:gradient_oth}
\end{figure*}

\begin{figure*}[!t]
    \centering
    \includegraphics[width=\linewidth]{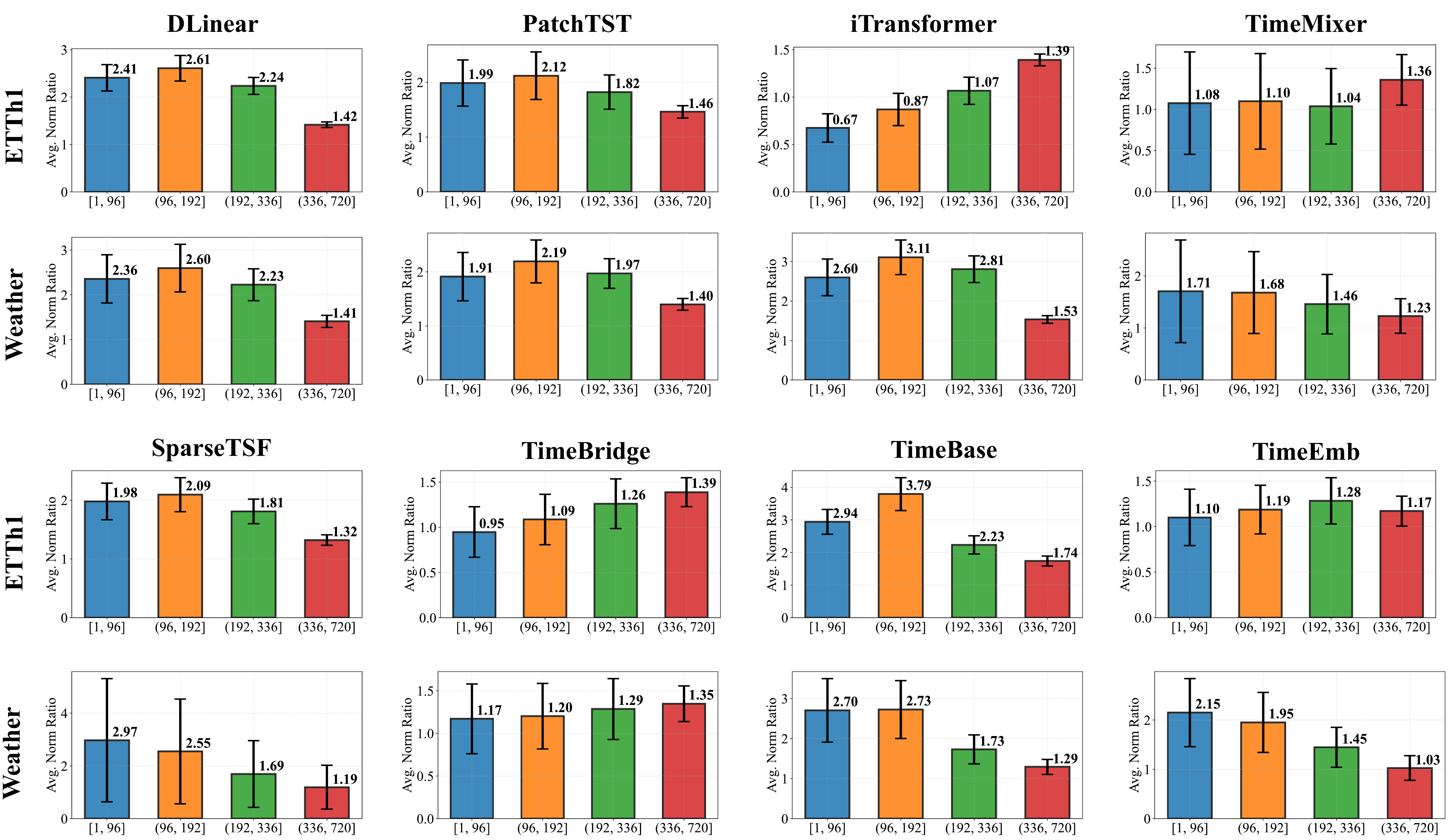}
    \caption{Visualization of the norm ratio between partitioned forecasting segments ($[1, 96]$ to $(336, 720]$) and the total sequence (``All") in DF paradigm on $T=L=720$ on ETTh1 and Weather through all training epochs.}
    \label{fig:gradient_norm}
\end{figure*}

We conducted experiments on two representative datasets: \textit{ETTh1} and \textit{Weather}. We report the statistics aggregated over the training process using two distinct visualization strategies:

\textbf{Global Static Analysis (Figure~\ref{fig:gradient_oth} \& Figure~\ref{fig:gradient_norm}).} 
We compute the global statistics over all batches and epochs to summarize the overall optimization landscape.
\begin{itemize}
    \item \textbf{Gradient Direction (Heatmap):} Figure~\ref{fig:gradient_oth} visualizes the pairwise cosine similarity matrix $\bar{\mathbf{M}}^{\text{Sim}}$ as a heatmap. Since the matrix is symmetric, we have retained only its upper triangular part for easier viewing. It reports the expectation of similarities between all segment pairs (including $s_{\text{all}}$) to reveal the conflict structure.
    \begin{equation}
        \bar{\mathbf{M}}_{i,j}^{\text{Sim}} = \frac{1}{E \cdot B} \sum_{e=1}^E \sum_{b=1}^B \text{Sim}_{i,j}^{(b,e)}.
    \end{equation}
    
    \item \textbf{Gradient Norm Ratio (Bar Chart):} Figure~\ref{fig:gradient_norm} presents the Gradient Norm Ratio $R_s$ for the four partitioned segments ($s \in \{s_{96}, s_{192}, s_{336}, s_{720}\}$) as a bar chart. Note that $s_{\text{all}}$ is excluded as it serves as the normalization denominator ($R_{s_{\text{all}}} \equiv 1$). We report both the global mean $\mu_s^{\text{global}}$ and the global standard deviation $\sigma_s^{\text{global}}$ (error bars) to demonstrate the variance of gradient dominance:
    \begin{equation}
        \mu_s^{\text{global}} = \mathbb{E}_{e,b}[R_s^{(b,e)}], \quad \sigma_s^{\text{global}} = \sqrt{\mathbb{E}_{e,b}\left[\left(R_s^{(b,e)} - \mu_s^{\text{global}}\right)^2\right]}.
    \end{equation}
\end{itemize}

\textbf{Training Dynamics (Figure~\ref{fig:gradient_oth_epoch} \& Figure~\ref{fig:gradient_norm_epoch}).} 
To trace the pathology evolution throughout training, we employ line charts with error bands to visualize how the relationship between each segment $s$ and the total update direction $s_{\text{all}}$ changes over time.
\begin{itemize}
    \item \textbf{Direction Consistency (Figure~\ref{fig:gradient_oth_epoch}):} Tracks the Cosine Similarity $\text{Sim}_{s, \text{all}}$.
    \item \textbf{Norm Contribution (Figure~\ref{fig:gradient_norm_epoch}):} Tracks the Norm Ratio $R_s$.
\end{itemize}
For both metrics, we compute the intra-epoch statistics. For a specific epoch $e$ and segment $s$, we report the mean $\mu_s^{(e)}$ and standard deviation $\sigma_s^{(e)}$ over all batches $B$ within that epoch:
\begin{equation}
    \mu_s^{(e)} = \frac{1}{B} \sum_{b=1}^B \mathcal{M}_{s}^{(b,e)}, \quad \sigma_s^{(e)} = \sqrt{\frac{1}{B} \sum_{b=1}^B \left( \mathcal{M}_{s}^{(b,e)} - \mu_s^{(e)} \right)^2},
\end{equation}
where $\mathcal{M}$ represents either the Similarity metric $\text{Sim}_{s, \text{all}}$ or the Ratio metric $R_s$.

\subsubsection{Analysis Results: Unveiling the Optimization Pathology}

Based on the visualizations, we present a detailed dissection of the optimization pathology in Direct Forecasting (DF).

\begin{figure*}[!t]
    \centering
    \includegraphics[width=\linewidth]{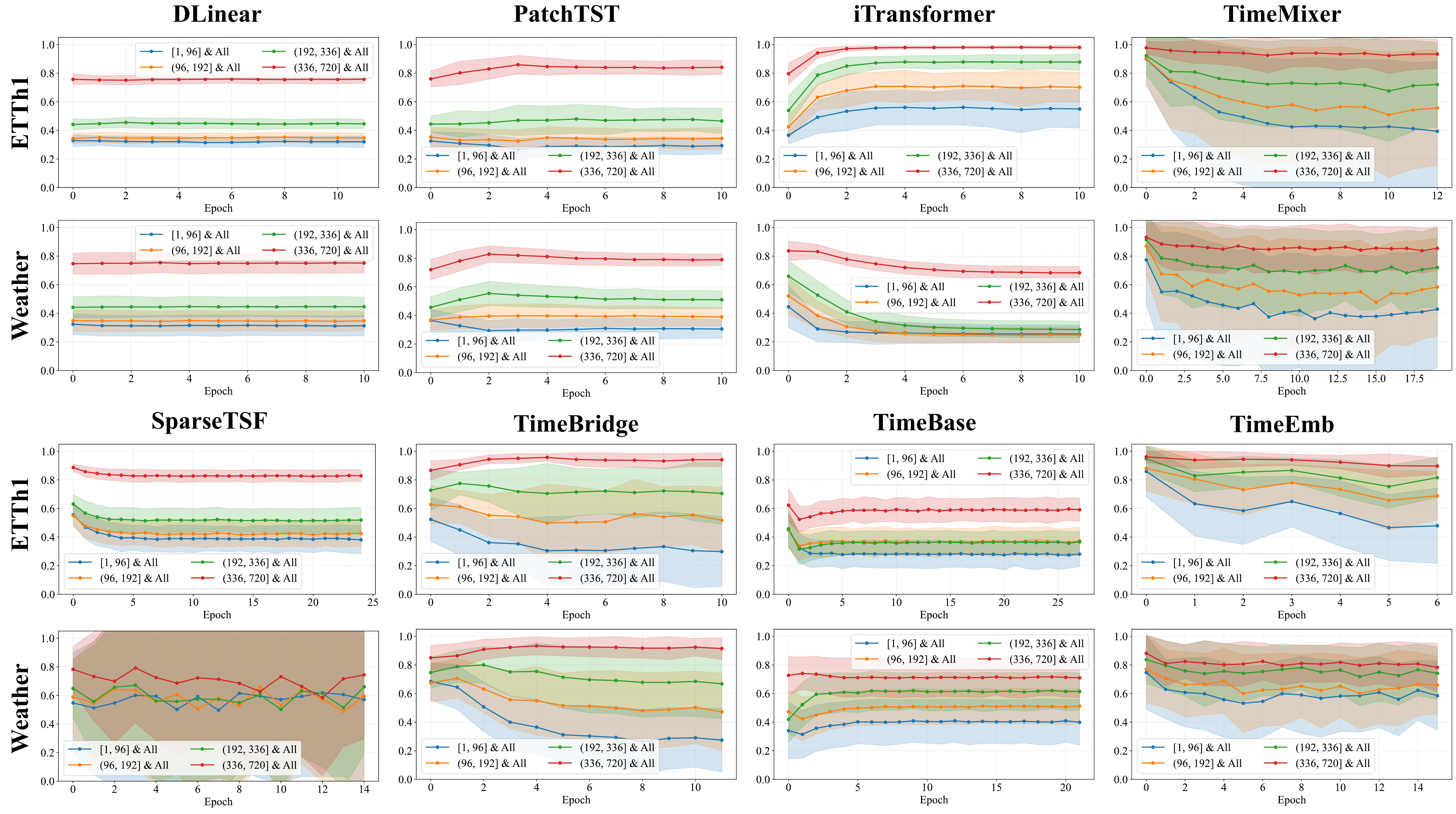}
    \caption{Visualization of Gradient Conflict and Distal Dominance in DF paradigm on $T=L=720$. We compute the average cosine similarity of gradients across partitioned forecasting segments ($[1, 96]$ to $(336, 720]$) and the total sequence (``All") during the training of all baselines on ETTh1 and Weather through all training epochs.}
    \label{fig:gradient_oth_epoch}
\end{figure*}

\textbf{Gradient Direction Analysis.} Observations from the gradient direction heatmaps (Figure~\ref{fig:gradient_oth}) and training dynamics (Figure~\ref{fig:gradient_oth_epoch}) reveal two systemic anomalies: \ding{202} \textbf{Distal Alignment Bias (The Hijacking).} We observe a strict \textit{monotonic increase} in the cosine similarity between segment gradients and the global update direction ($s_{\text{all}}$) as the segment moves further into the future. This indicates that the final parameter update is predominantly governed by the optimization objectives of the distal horizon. As shown in the dynamic plots (Figure~\ref{fig:gradient_oth_epoch}), the furthest segment (Red Line) consistently maintains the highest similarity with $s_{\text{all}}$ throughout the entire training process. \ding{203} \textbf{Gradient Conflict.} The similarity between segments decreases monotonically as their relative temporal distance increases. Crucially, we observe severe \textit{orthogonality} or even \textit{opposition} between proximal and distal segments. For instance, in \textit{iTransformer} and \textit{PatchTST}, the cosine similarity between the nearest and furthest segments approaches zero or becomes negative. In the extreme case of \textit{iTransformer} on the \textit{Weather} dataset, the average similarity drops to $\sim 0.2$ ($\approx 78.5^\circ$), implying that the optimization direction required for the near-term is nearly orthogonal to the actual update direction. This confirms that long-horizon training induces severe internal gradient conflicts.

\textit{Remark on DLinear:} Note that for DLinear, the inter-segment similarity is strictly zero. This is an expected structural artifact, as DLinear employs independent linear weights for each time step, isolating their gradients. However, its interaction with the global update ($s_{\text{all}}$) still exhibits the same Distal Alignment Bias, corroborating the universality of this phenomenon.

\begin{figure*}[!t]
    \centering
    \includegraphics[width=\linewidth]{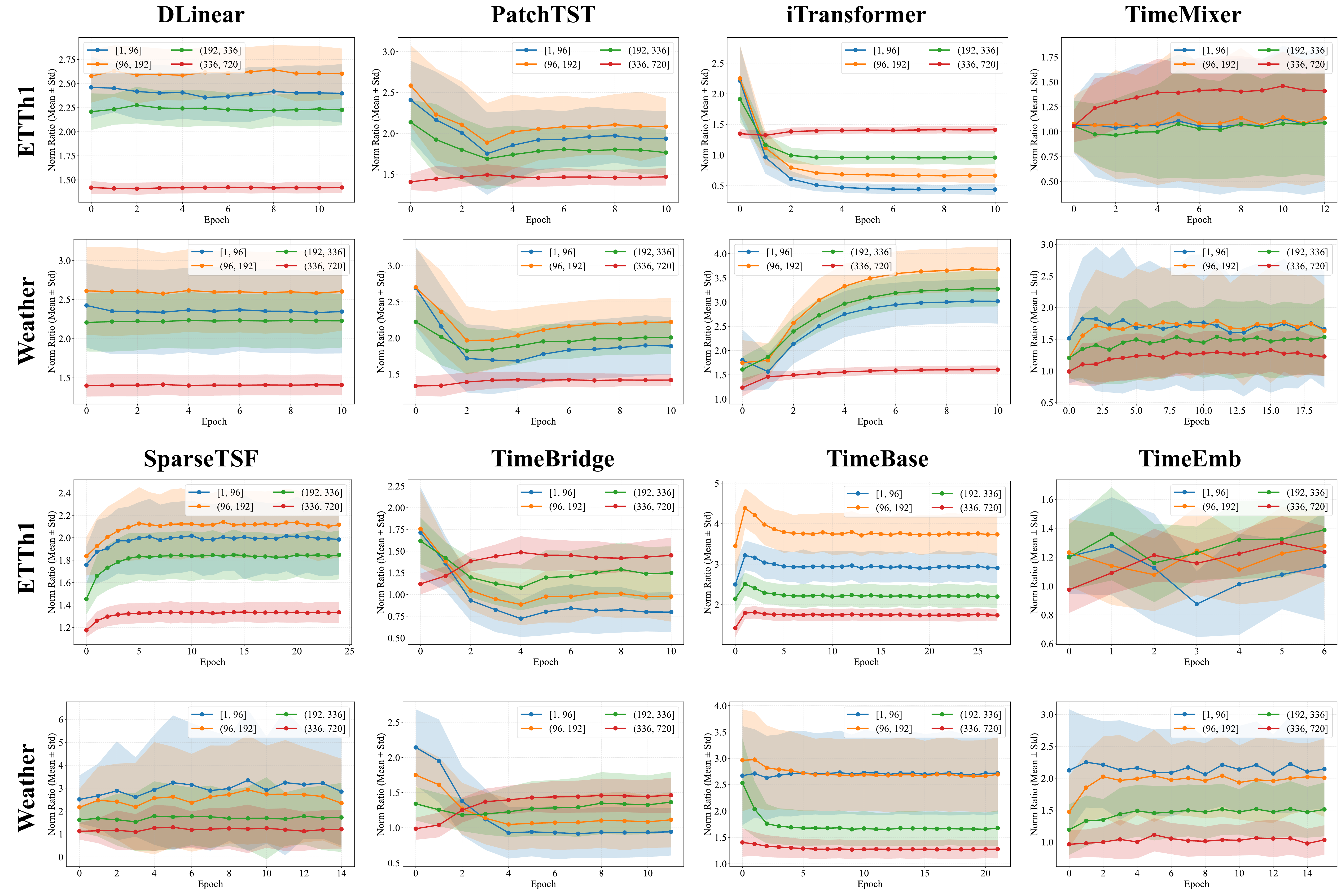}
    \caption{Visualization of the norm ratio between partitioned forecasting segments ($[1, 96]$ to $(336, 720]$) and the total sequence (``All") in DF paradigm on $T=L=720$ on ETTh1 and Weather through each training epoch.}
    \label{fig:gradient_norm_epoch}
\end{figure*}

\textbf{Gradient Norm Analysis.}
A natural question arises: \textbf{Does the model favor the distal horizon because the proximal horizon is already well-optimized?} To answer this, we analyze the gradient norms (Figure~\ref{fig:gradient_norm} \& Figure~\ref{fig:gradient_norm_epoch}), where a larger norm implies a higher demand for parameter adaptation (i.e., the model is far from the local optimum). \ding{202} \textbf{Proximal High-Norm Phenomenon.} Contrary to the ``well-optimized'' hypothesis, we find that the gradient norms are actually largest in the nearest segments (e.g., $[1, 96]$) and smallest in the distal segments for the vast majority of models. This reveals a startling Optimization Paradox: the near-term segments generate the strongest error signals (high norms) indicating a need for learning, yet the global update direction is hijacked by the distal segments (which have smaller norms). \ding{203} \textbf{Dynamics and Exceptions.} While iTransformer and TimeMixer on ETTh1 initially show larger distal norms, a closer inspection of the epoch dynamics (Figure~\ref{fig:gradient_norm_epoch}) reveals that during the effective training phase (before early-stopping convergence), the proximal norms are indeed dominant or competitive. The late-stage dominance of distal norms in these specific cases likely reflects noise overfitting rather than valid learning. In general, the pattern of ``High Proximal Norm vs. Low Distal Norm'' holds consistently.

Synthesizing the Direction and Norm analyses, we identify the root cause of the performance degradation in long-horizon DF: Although the proximal horizon exhibits the most significant underfitting (High Norms), the conflicting gradient directions from the distal horizon forcefully steer the optimization trajectory away from the proximal needs. Consequently, the model fails to capture near-term dynamics effectively. This Near-term Underfitting explains the counter-intuitive truncation error observed in Figure~\ref{fig:first_truncated}. By decoupling the training horizon, EF eliminates this conflict, allowing the model to focus its capacity on valid, high-fidelity signals (``Look Close''), thereby achieving superior generalization (``See Far'').

\subsection{Analysis of Extreme Extrapolation Stability}\label{sec:ees_analysis}Building upon the gradient pathology revealed in Appendix~\ref{app:optpath}, we provide a deeper theoretical justification for the superior stability of EF in extreme extrapolation regimes (e.g., $H=720$ or larger) in Section~\ref{sec:ees}. We attribute the catastrophic failure of ultra-long DF not merely to capacity limitations, but to a Dual-Dilemma unique to extreme LTSF: \ding{202} \textbf{Severe Optimization Pathology (Gradient Perspective).} As rigorously proven in Appendix~\ref{app:optpath}, expanding the training horizon $L$ to match a massive $H$ ($L=H$) exponentially exacerbates optimization difficulties. The Distal Hijacking effect forces the model to prioritize distant, high-uncertainty targets at the expense of capturing near-term dynamics, leading to a breakdown in learning valid temporal patterns. \ding{203} \textbf{Sample Impoverishment Bottleneck (Data Perspective).} Beyond optimization, we identify a structural vulnerability in the DF data pipeline. For a time series dataset of total length $\mathcal{N}$, the number of valid training samples $N_{\text{samples}}$ generated via a sliding window mechanism with input length $T$ and output horizon $L$ is strictly defined as:
\begin{equation}N_{\text{samples}}(L) = \max\left(0, \mathcal{N} - (T + L) + 1\right).
\end{equation}This formulation highlights a critical inverse relationship:
\begin{itemize}
    \item \textbf{For DF ($L=H_{\text{extreme}}$):} As the target horizon $H_{\text{extreme}}$ grows significantly, the term $(T + H_{\text{extreme}})$ approaches $\mathcal{N}$, causing $N_{\text{samples}}$ to plummet. In extreme cases where $T + H_{\text{extreme}} \approx \mathcal{N}$, the effective sample size collapses to $N_{\text{samples}} \to 1$. Even with batch augmentation strategies (e.g., duplicating samples to fill a batch~\cite{qiu2024tfb}), the Effective Information Capacity is restricted to a single data, leading to poor generalization.

    \item \textbf{For EF ($L\ll H_{\text{extreme}}$):} EF maintains a fixed, optimal reasoning horizon $L \ll H_{\text{extreme}}$. Consequently, the available sample size $N_{\text{samples}}(L_{\text{opt}})$ remains constant and abundant regardless of the inference requirement $H_{\text{extreme}}$.
\end{itemize}

This analysis culminates in a profound conclusion regarding the stability of extreme forecasting. While EF introduces a theoretical Recursive Accumulation Error, our empirical and theoretical evidence confirms that this error is significantly less detrimental than the compound impact of optimization pathology and Sample Impoverishment that cripples the DF paradigm. By decoupling the training horizon from the inference task, EF ensures both healthy gradient dynamics and data abundance ("Look Close"), thereby sustaining robust performance even in extreme extrapolation ("See Far").

\begin{table*}[!t]
  \centering
  \caption{Performance landscape of \textbf{DLinear} on the $(T, L, H)$ space. $T,L,H$ are the input length, output horizon and evaluation horizon, respectively. Results under the conventional DF paradigm ($L=H$) are highlighted in \colorbox[HTML]{F6E9C5}{Orange Background}. For each specific $T$ and $H$, the best performance is \underline{\textcolor{blue}{Underlined}}, while the global optimum across all input length for a fixed $H$ is denoted in \textbf{\textbf{\textcolor{red}{Bold}}}.}
  \aboverulesep=0pt
  \belowrulesep=0pt
  \setlength{\arrayrulewidth}{1.0pt}
  \setlength{\tabcolsep}{2.5pt}
  \resizebox{\linewidth}{!}{
\begin{tabular}{ccc|cc|cc|cc|cc|ccc|cc|cc|cc|cc}
    \toprule
    \multicolumn{22}{c}{\textbf{DLinear}} \\
    \midrule
    \multirow{2}[2]{*}{} & \multirow{2}[2]{*}{{$T$}} & \multirow{2}[2]{*}{\diagbox{$L$}{$H$}} & \multicolumn{2}{c|}{{$96$}} & \multicolumn{2}{c|}{{$192$}} & \multicolumn{2}{c|}{{$336$}} & \multicolumn{2}{c|}{{$720$}} & \multirow{2}[2]{*}{} & \multirow{2}[2]{*}{{$T$}} & \multirow{2}[2]{*}{\diagbox{$L$}{$H$}} & \multicolumn{2}{c|}{{$96$}} & \multicolumn{2}{c|}{{$192$}} & \multicolumn{2}{c|}{{$336$}} & \multicolumn{2}{c}{{$720$}} \\
          &       &       & {MSE} & {MAE} & {MSE} & {MAE} & {MSE} & {MAE} & {MSE} & {MAE} &       &       &       & {MSE} & {MAE} & {MSE} & {MAE} & {MSE} & {MAE} & {MSE} & {MAE} \\
    \midrule
    \midrule
    \multirow{16}[0]{*}{\begin{sideways}\textbf{ETTh1}\end{sideways}} & \multirow{4}[2]{*}{{$96$}} & {$96$} & \cellcolor[HTML]{F6E9C5}\textcolor[rgb]{ 0,  0,  1}{\underline{0.396}} & \cellcolor[HTML]{F6E9C5}\textcolor[rgb]{ 0,  0,  1}{\underline{0.410}} & 0.449 & 0.445 & 0.503 & 0.481 & 0.580 & 0.552 & \multirow{16}[0]{*}{\begin{sideways}\textbf{Weather}\end{sideways}} & \multirow{4}[2]{*}{{$96$}} & {$96$} & \cellcolor[HTML]{F6E9C5}0.196 & \cellcolor[HTML]{F6E9C5}\textcolor[rgb]{ 0,  0,  1}{\underline{0.255}} & 0.237 & 0.297 & 0.288 & 0.344 & 0.379 & 0.422 \\
          &       & {$192$} & \textcolor[rgb]{ 0,  0,  1}{\underline{0.396}} & 0.411 & \cellcolor[HTML]{F6E9C5}\textcolor[rgb]{ 0,  0,  1}{\underline{0.445}} & \cellcolor[HTML]{F6E9C5}\textcolor[rgb]{ 0,  0,  1}{\underline{0.440}} & 0.492 & 0.472 & 0.544 & 0.531 &       &       & {$192$} & 0.200 & 0.264 & \cellcolor[HTML]{F6E9C5}0.239 & \cellcolor[HTML]{F6E9C5}0.299 & 0.290 & 0.346 & 0.380 & 0.423 \\
\bottomrule
    \end{tabular}}%
  \label{tab:dlinear_all}%
\end{table*}%

\begin{table*}[!t]
  \centering
  \caption{Performance landscape of \textbf{PatchTST} on the $(T, L, H)$ space. $T,L,H$ are the input length, output horizon and evaluation horizon, respectively. Results under the conventional DF paradigm ($L=H$) are highlighted in \colorbox[HTML]{F6E9C5}{Orange Background}. For each specific $T$ and $H$, the best performance is \underline{\textcolor{blue}{Underlined}}, while the global optimum across all input length for a fixed $H$ is denoted in \textbf{\textcolor{red}{Bold}}.}
  \aboverulesep=0pt
  \belowrulesep=0pt
  \setlength{\arrayrulewidth}{1.0pt}
  \setlength{\tabcolsep}{2.5pt}
  \resizebox{\linewidth}{!}{
\begin{tabular}{ccc|cc|cc|cc|cc|ccc|cc|cc|cc|cc}
    \toprule
    \multicolumn{22}{c}{\textbf{PatchTST}} \\
    \midrule
    \multirow{2}[2]{*}{} & \multirow{2}[2]{*}{{$T$}} & \multirow{2}[2]{*}{\diagbox{$L$}{$H$}} & \multicolumn{2}{c|}{{$96$}} & \multicolumn{2}{c|}{{$192$}} & \multicolumn{2}{c|}{{$336$}} & \multicolumn{2}{c|}{{$720$}} & \multirow{2}[2]{*}{} & \multirow{2}[2]{*}{{$T$}} & \multirow{2}[2]{*}{\diagbox{$L$}{$H$}} & \multicolumn{2}{c|}{{$96$}} & \multicolumn{2}{c|}{{$192$}} & \multicolumn{2}{c|}{{$336$}} & \multicolumn{2}{c}{{$720$}} \\
          &       &       & {MSE} & {MAE} & {MSE} & {MAE} & {MSE} & {MAE} & {MSE} & {MAE} &       &       &       & {MSE} & {MAE} & {MSE} & {MAE} & {MSE} & {MAE} & {MSE} & {MAE} \\
    \midrule
    \midrule
    \multirow{16}[0]{*}{\begin{sideways}\textbf{ETTh1}\end{sideways}} & \multirow{4}[2]{*}{{$96$}} & {$96$} & \cellcolor[HTML]{F6E9C5}\textcolor[rgb]{ 0,  0,  1}{\underline{0.380}} & \cellcolor[HTML]{F6E9C5}\textcolor[rgb]{ 1,  0,  0}{\textbf{0.402}} & 0.429 & \textcolor[rgb]{ 0,  0,  1}{\underline{0.432}} & 0.478 & 0.456 & 0.532 & 0.483 & \multirow{16}[0]{*}{\begin{sideways}\textbf{Weather}\end{sideways}} & \multirow{4}[2]{*}{{$96$}} & {$96$} & \cellcolor[HTML]{F6E9C5}\textcolor[rgb]{ 0,  0,  1}{\underline{0.172}} & \cellcolor[HTML]{F6E9C5}\textcolor[rgb]{ 0,  0,  1}{\underline{0.213}} & 0.222 & 0.257 & 0.277 & 0.298 & 0.358 & 0.351 \\
\bottomrule
    \end{tabular}}%
  \label{tab:patchtst_all}%
\end{table*}%

\begin{table*}[!t]
  \centering
  \caption{Performance landscape of \textbf{iTransformer} on the $(T, L, H)$ space. $T,L,H$ are the input length, output horizon and evaluation horizon, respectively. Results under the conventional DF paradigm ($L=H$) are highlighted in \colorbox[HTML]{F6E9C5}{Orange Background}. For each specific $T$ and $H$, the best performance is \underline{\textcolor{blue}{Underlined}}, while the global optimum across all input length for a fixed $H$ is denoted in \textbf{\textcolor{red}{Bold}}.}
  \aboverulesep=0pt
  \belowrulesep=0pt
  \setlength{\arrayrulewidth}{1.0pt}
  \setlength{\tabcolsep}{2.5pt}
  \resizebox{\linewidth}{!}{
\begin{tabular}{ccc|cc|cc|cc|cc|ccc|cc|cc|cc|cc}
    \toprule
    \multicolumn{22}{c}{\textbf{iTransformer}} \\
    \midrule
    \multirow{2}[2]{*}{} & \multirow{2}[2]{*}{{$T$}} & \multirow{2}[2]{*}{\diagbox{$L$}{$H$}} & \multicolumn{2}{c|}{{$96$}} & \multicolumn{2}{c|}{{$192$}} & \multicolumn{2}{c|}{{$336$}} & \multicolumn{2}{c|}{{$720$}} & \multirow{2}[2]{*}{} & \multirow{2}[2]{*}{{$T$}} & \multirow{2}[2]{*}{\diagbox{$L$}{$H$}} & \multicolumn{2}{c|}{{$96$}} & \multicolumn{2}{c|}{{$192$}} & \multicolumn{2}{c|}{{$336$}} & \multicolumn{2}{c}{{$720$}} \\
          &       &       & {MSE} & {MAE} & {MSE} & {MAE} & {MSE} & {MAE} & {MSE} & {MAE} &       &       &       & {MSE} & {MAE} & {MSE} & {MAE} & {MSE} & {MAE} & {MSE} & {MAE} \\
    \midrule
    \midrule
    \multirow{16}[0]{*}{\begin{sideways}\textbf{ETTh1}\end{sideways}} & \multirow{4}[2]{*}{{$96$}} & {$96$} & \cellcolor[HTML]{F6E9C5}\textcolor[rgb]{ 0,  0,  1}{\underline{0.395}} & \cellcolor[HTML]{F6E9C5}\textcolor[rgb]{ 1,  0,  0}{\textbf{0.409}} & \textcolor[rgb]{ 0,  0,  1}{\underline{0.448}} & 0.442 & 0.498 & 0.472 & 0.541 & 0.515 & \multirow{16}[0]{*}{\begin{sideways}\textbf{Weather}\end{sideways}} & \multirow{4}[2]{*}{{$96$}} & {$96$} & \cellcolor[HTML]{F6E9C5}\textcolor[rgb]{ 0,  0,  1}{\underline{0.175}} & \cellcolor[HTML]{F6E9C5}\textcolor[rgb]{ 0,  0,  1}{\underline{0.216}} & 0.225 & 0.259 & 0.281 & 0.299 & 0.359 & 0.350 \\
\bottomrule
    \end{tabular}}%
  \label{tab:itransformer_all}%
\end{table*}%

\begin{table*}[!t]
  \centering
  \caption{Performance landscape of \textbf{TimeMixer} on the $(T, L, H)$ space. $T,L,H$ are the input length, output horizon and evaluation horizon, respectively. Results under the conventional DF paradigm ($L=H$) are highlighted in \colorbox[HTML]{F6E9C5}{Orange Background}. For each specific $T$ and $H$, the best performance is \underline{\textcolor{blue}{Underlined}}, while the global optimum across all input length for a fixed $H$ is denoted in \textbf{\textcolor{red}{Bold}}.}
  \aboverulesep=0pt
  \belowrulesep=0pt
  \setlength{\arrayrulewidth}{1.0pt}
  \setlength{\tabcolsep}{2.5pt}
  \resizebox{\linewidth}{!}{
\begin{tabular}{ccc|cc|cc|cc|cc|ccc|cc|cc|cc|cc}
    \toprule
    \multicolumn{22}{c}{\textbf{TimeMixer}} \\
    \midrule
    \multirow{2}[2]{*}{} & \multirow{2}[2]{*}{{$T$}} & \multirow{2}[2]{*}{\diagbox{$L$}{$H$}} & \multicolumn{2}{c|}{{$96$}} & \multicolumn{2}{c|}{{$192$}} & \multicolumn{2}{c|}{{$336$}} & \multicolumn{2}{c|}{{$720$}} & \multirow{2}[2]{*}{} & \multirow{2}[2]{*}{{$T$}} & \multirow{2}[2]{*}{\diagbox{$L$}{$H$}} & \multicolumn{2}{c|}{{$96$}} & \multicolumn{2}{c|}{{$192$}} & \multicolumn{2}{c|}{{$336$}} & \multicolumn{2}{c}{{$720$}} \\
          &       &       & {MSE} & {MAE} & {MSE} & {MAE} & {MSE} & {MAE} & {MSE} & {MAE} &       &       &       & {MSE} & {MAE} & {MSE} & {MAE} & {MSE} & {MAE} & {MSE} & {MAE} \\
    \midrule
    \midrule
    \multirow{16}[0]{*}{\begin{sideways}\textbf{ETTh1}\end{sideways}} & \multirow{4}[2]{*}{{$96$}} & {$96$} & \cellcolor[HTML]{F6E9C5}\textcolor[rgb]{ 0,  0,  1}{\underline{0.379}} & \cellcolor[HTML]{F6E9C5}\textcolor[rgb]{ 1,  0,  0}{\textbf{0.396}} & \textcolor[rgb]{ 0,  0,  1}{\underline{0.431}} & 0.429 & \textcolor[rgb]{ 0,  0,  1}{\underline{0.476}} & 0.454 & 0.505 & 0.480 & \multirow{16}[0]{*}{\begin{sideways}\textbf{Weather}\end{sideways}} & \multirow{4}[2]{*}{{$96$}} & {$96$} & \cellcolor[HTML]{F6E9C5}\textcolor[rgb]{ 0,  0,  1}{\underline{0.162}} & \cellcolor[HTML]{F6E9C5}\textcolor[rgb]{ 0,  0,  1}{\underline{0.209}} & 0.209 & \textcolor[rgb]{ 0,  0,  1}{\underline{0.251}} & 0.264 & \textcolor[rgb]{ 0,  0,  1}{\underline{0.291}} & 0.347 & 0.343 \\
\bottomrule
    \end{tabular}}%
  \label{tab:timemier_all}%
\end{table*}%

\begin{table*}[!t]
  \centering
  \caption{Performance landscape of \textbf{SparseTSF} on the $(T, L, H)$ space. $T,L,H$ are the input length, output horizon and evaluation horizon, respectively. Results under the conventional DF paradigm ($L=H$) are highlighted in \colorbox[HTML]{F6E9C5}{Orange Background}. For each specific $T$ and $H$, the best performance is \underline{\textcolor{blue}{Underlined}}, while the global optimum across all input length for a fixed $H$ is denoted in \textbf{\textcolor{red}{Bold}}.}
  \aboverulesep=0pt
  \belowrulesep=0pt
  \setlength{\arrayrulewidth}{1.0pt}
  \setlength{\tabcolsep}{2.5pt}
  \resizebox{\linewidth}{!}{
\begin{tabular}{ccc|cc|cc|cc|cc|ccc|cc|cc|cc|cc}
    \toprule
    \multicolumn{22}{c}{\textbf{SparseTSF}} \\
    \midrule
    \multirow{2}[2]{*}{} & \multirow{2}[2]{*}{{$T$}} & \multirow{2}[2]{*}{\diagbox{$L$}{$H$}} & \multicolumn{2}{c|}{{$96$}} & \multicolumn{2}{c|}{{$192$}} & \multicolumn{2}{c|}{{$336$}} & \multicolumn{2}{c|}{{$720$}} & \multirow{2}[2]{*}{} & \multirow{2}[2]{*}{{$T$}} & \multirow{2}[2]{*}{\diagbox{$L$}{$H$}} & \multicolumn{2}{c|}{{$96$}} & \multicolumn{2}{c|}{{$192$}} & \multicolumn{2}{c|}{{$336$}} & \multicolumn{2}{c}{{$720$}} \\
          &       &       & {MSE} & {MAE} & {MSE} & {MAE} & {MSE} & {MAE} & {MSE} & {MAE} &       &       &       & {MSE} & {MAE} & {MSE} & {MAE} & {MSE} & {MAE} & {MSE} & {MAE} \\
    \midrule
    \midrule
    \multirow{16}[0]{*}{\begin{sideways}\textbf{ETTh1}\end{sideways}} & \multirow{4}[2]{*}{{$96$}} & {$96$} & \cellcolor[HTML]{F6E9C5}\textcolor[rgb]{ 0,  0,  1}{\underline{0.581}} & \cellcolor[HTML]{F6E9C5}\textcolor[rgb]{ 0,  0,  1}{\underline{0.510}} & 0.633 & 0.535 & 0.675 & 0.559 & 0.693 & 0.585 & \multirow{16}[0]{*}{\begin{sideways}\textbf{Weather}\end{sideways}} & \multirow{4}[2]{*}{{$96$}} & {$96$} & \cellcolor[HTML]{F6E9C5}0.198 & \cellcolor[HTML]{F6E9C5}\textcolor[rgb]{ 0,  0,  1}{\underline{0.237}} & 0.243 & 0.275 & 0.294 & 0.312 & 0.368 & 0.362 \\
          &       & {$192$} & \textcolor[rgb]{ 0,  0,  1}{\underline{0.581}} & \textcolor[rgb]{ 0,  0,  1}{\underline{0.510}} & \cellcolor[HTML]{F6E9C5}\textcolor[rgb]{ 0,  0,  1}{\underline{0.602}} & \cellcolor[HTML]{F6E9C5}\textcolor[rgb]{ 0,  0,  1}{\underline{0.524}} & 0.638 & 0.544 & 0.663 & 0.572 &       &       & {$192$} & 0.197 & \textcolor[rgb]{ 0,  0,  1}{\underline{0.237}} & \cellcolor[HTML]{F6E9C5}\textcolor[rgb]{ 0,  0,  1}{\underline{0.242}} & \cellcolor[HTML]{F6E9C5}0.273 & \textcolor[rgb]{ 0,  0,  1}{\underline{0.293}} & 0.311 & 0.366 & 0.360 \\
\bottomrule
    \end{tabular}}%
  \label{tab:sparsetsf_all}%
\end{table*}%

\begin{table*}[!t]
  \centering
  \caption{Performance landscape of \textbf{TimeBridge} on the $(T, L, H)$ space. $T,L,H$ are the input length, output horizon and evaluation horizon, respectively. Results under the conventional DF paradigm ($L=H$) are highlighted in \colorbox[HTML]{F6E9C5}{Orange Background}. For each specific $T$ and $H$, the best performance is \underline{\textcolor{blue}{Underlined}}, while the global optimum across all input length for a fixed $H$ is denoted in \textbf{\textcolor{red}{Bold}}.}
  \aboverulesep=0pt
  \belowrulesep=0pt
  \setlength{\arrayrulewidth}{1.0pt}
  \setlength{\tabcolsep}{2.5pt}
  \resizebox{\linewidth}{!}{
\begin{tabular}{ccc|cc|cc|cc|cc|ccc|cc|cc|cc|cc}
    \toprule
    \multicolumn{22}{c}{\textbf{TimeBridge}} \\
    \midrule
    \multirow{2}[2]{*}{} & \multirow{2}[2]{*}{{$T$}} & \multirow{2}[2]{*}{\diagbox{$L$}{$H$}} & \multicolumn{2}{c|}{{$96$}} & \multicolumn{2}{c|}{{$192$}} & \multicolumn{2}{c|}{{$336$}} & \multicolumn{2}{c|}{{$720$}} & \multirow{2}[2]{*}{} & \multirow{2}[2]{*}{{$T$}} & \multirow{2}[2]{*}{\diagbox{$L$}{$H$}} & \multicolumn{2}{c|}{{$96$}} & \multicolumn{2}{c|}{{$192$}} & \multicolumn{2}{c|}{{$336$}} & \multicolumn{2}{c}{{$720$}} \\
          &       &       & {MSE} & {MAE} & {MSE} & {MAE} & {MSE} & {MAE} & {MSE} & {MAE} &       &       &       & {MSE} & {MAE} & {MSE} & {MAE} & {MSE} & {MAE} & {MSE} & {MAE} \\
    \midrule
    \midrule
    \multirow{16}[0]{*}{\begin{sideways}\textbf{ETTh1}\end{sideways}} & \multirow{4}[2]{*}{{$96$}} & {$96$} & \cellcolor[HTML]{F6E9C5}\textcolor[rgb]{ 0,  0,  1}{\underline{0.380}} & \cellcolor[HTML]{F6E9C5}\textcolor[rgb]{ 1,  0,  0}{\textbf{0.398}} & \textcolor[rgb]{ 0,  0,  1}{\underline{0.426}} & \textcolor[rgb]{ 0,  0,  1}{\underline{0.428}} & 0.466 & 0.450 & 0.484 & 0.476 & \multirow{16}[0]{*}{\begin{sideways}\textbf{Weather}\end{sideways}} & \multirow{4}[2]{*}{{$96$}} & {$96$} & \cellcolor[HTML]{F6E9C5}\textcolor[rgb]{ 0,  0,  1}{\underline{0.181}} & \cellcolor[HTML]{F6E9C5}\textcolor[rgb]{ 0,  0,  1}{\underline{0.221}} & 0.229 & \textcolor[rgb]{ 0,  0,  1}{\underline{0.261}} & 0.284 & 0.300 & 0.362 & 0.351 \\
\bottomrule
    \end{tabular}}%
  \label{tab:timebridge_all}%
\end{table*}%

\begin{table*}[!t]
  \centering
  \caption{Performance landscape of \textbf{TimeBase} on the $(T, L, H)$ space. $T,L,H$ are the input length, output horizon and evaluation horizon, respectively. Results under the conventional DF paradigm ($L=H$) are highlighted in \colorbox[HTML]{F6E9C5}{Orange Background}. For each specific $T$ and $H$, the best performance is \underline{\textcolor{blue}{Underlined}}, while the global optimum across all input length for a fixed $H$ is denoted in \textbf{\textcolor{red}{Bold}}.}
  \aboverulesep=0pt
  \belowrulesep=0pt
  \setlength{\arrayrulewidth}{1.0pt}
  \setlength{\tabcolsep}{2.5pt}
  \resizebox{\linewidth}{!}{
\begin{tabular}{ccc|cc|cc|cc|cc|ccc|cc|cc|cc|cc}
    \toprule
    \multicolumn{22}{c}{\textbf{TimeBase}} \\
    \midrule
    \multirow{2}[2]{*}{} & \multirow{2}[2]{*}{{$T$}} & \multirow{2}[2]{*}{\diagbox{$L$}{$H$}} & \multicolumn{2}{c|}{{$96$}} & \multicolumn{2}{c|}{{$192$}} & \multicolumn{2}{c|}{{$336$}} & \multicolumn{2}{c|}{{$720$}} & \multirow{2}[2]{*}{} & \multirow{2}[2]{*}{{$T$}} & \multirow{2}[2]{*}{\diagbox{$L$}{$H$}} & \multicolumn{2}{c|}{{$96$}} & \multicolumn{2}{c|}{{$192$}} & \multicolumn{2}{c|}{{$336$}} & \multicolumn{2}{c}{{$720$}} \\
          &       &       & {MSE} & {MAE} & {MSE} & {MAE} & {MSE} & {MAE} & {MSE} & {MAE} &       &       &       & {MSE} & {MAE} & {MSE} & {MAE} & {MSE} & {MAE} & {MSE} & {MAE} \\
    \midrule
    \midrule
    \multirow{16}[0]{*}{\begin{sideways}\textbf{ETTh1}\end{sideways}} & \multirow{4}[2]{*}{{$96$}} & {$96$} & \cellcolor[HTML]{F6E9C5}\textcolor[rgb]{ 0,  0,  1}{\underline{0.632}} & \cellcolor[HTML]{F6E9C5}\textcolor[rgb]{ 0,  0,  1}{\underline{0.532}} & \textcolor[rgb]{ 0,  0,  1}{\underline{0.678}} & \textcolor[rgb]{ 0,  0,  1}{\underline{0.557}} & \textcolor[rgb]{ 0,  0,  1}{\underline{0.708}} & \textcolor[rgb]{ 0,  0,  1}{\underline{0.578}} & \textcolor[rgb]{ 0,  0,  1}{\underline{0.732}} & \textcolor[rgb]{ 0,  0,  1}{\underline{0.608}} & \multirow{16}[0]{*}{\begin{sideways}\textbf{Weather}\end{sideways}} & \multirow{4}[2]{*}{{$96$}} & {$96$} & \cellcolor[HTML]{F6E9C5}\textcolor[rgb]{ 0,  0,  1}{\underline{0.205}} & \cellcolor[HTML]{F6E9C5}\textcolor[rgb]{ 0,  0,  1}{\underline{0.249}} & \textcolor[rgb]{ 0,  0,  1}{\underline{0.249}} & \textcolor[rgb]{ 0,  0,  1}{\underline{0.284}} & \textcolor[rgb]{ 0,  0,  1}{\underline{0.299}} & 0.320 & 0.371 & 0.368 \\
\bottomrule
    \end{tabular}}%
  \label{tab:timebase_all}%
\end{table*}%

\begin{table*}[!t]
  \centering
  \caption{Performance landscape of \textbf{TimeEmb} on the $(T, L, H)$ space. $T,L,H$ are the input length, output horizon and evaluation horizon, respectively. Results under the conventional DF paradigm ($L=H$) are highlighted in \colorbox[HTML]{F6E9C5}{Orange Background}. For each specific $T$ and $H$, the best performance is \underline{\textcolor{blue}{Underlined}}, while the global optimum across all input length for a fixed $H$ is denoted in \textbf{\textcolor{red}{Bold}}.}
  \aboverulesep=0pt
  \belowrulesep=0pt
  \setlength{\arrayrulewidth}{1.0pt}
  \setlength{\tabcolsep}{2.5pt}
  \resizebox{\linewidth}{!}{
\begin{tabular}{ccc|cc|cc|cc|cc|ccc|cc|cc|cc|cc}
    \toprule
    \multicolumn{22}{c}{\textbf{TimeEmb}} \\
    \midrule
    \multirow{2}[2]{*}{} & \multirow{2}[2]{*}{{$T$}} & \multirow{2}[2]{*}{\diagbox{$L$}{$H$}} & \multicolumn{2}{c|}{{$96$}} & \multicolumn{2}{c|}{{$192$}} & \multicolumn{2}{c|}{{$336$}} & \multicolumn{2}{c|}{{$720$}} & \multirow{2}[2]{*}{} & \multirow{2}[2]{*}{{$T$}} & \multirow{2}[2]{*}{\diagbox{$L$}{$H$}} & \multicolumn{2}{c|}{{$96$}} & \multicolumn{2}{c|}{{$192$}} & \multicolumn{2}{c|}{{$336$}} & \multicolumn{2}{c}{{$720$}} \\
          &       &       & {MSE} & {MAE} & {MSE} & {MAE} & {MSE} & {MAE} & {MSE} & {MAE} &       &       &       & {MSE} & {MAE} & {MSE} & {MAE} & {MSE} & {MAE} & {MSE} & {MAE} \\
    \midrule
    \midrule
    \multirow{16}[0]{*}{\begin{sideways}\textbf{ETTh1}\end{sideways}} & \multirow{4}[2]{*}{{$96$}} & {$96$} & \cellcolor[HTML]{F6E9C5}0.387 & \cellcolor[HTML]{F6E9C5}0.400 & 0.436 & 0.429 & 0.471 & 0.448 & 0.459 & 0.459 & \multirow{16}[0]{*}{\begin{sideways}\textbf{Weather}\end{sideways}} & \multirow{4}[2]{*}{{$96$}} & {$96$} & \cellcolor[HTML]{F6E9C5}\textcolor[rgb]{ 0,  0,  1}{\underline{0.155}} & \cellcolor[HTML]{F6E9C5}\textcolor[rgb]{ 1,  0,  0}{\textbf{0.199}} & 0.205 & \textcolor[rgb]{ 0,  0,  1}{\underline{0.243}} & 0.262 & \textcolor[rgb]{ 0,  0,  1}{\underline{0.285}} & 0.346 & 0.338 \\
\bottomrule
    \end{tabular}}%
  \label{tab:timeemb_all}%
\end{table*}%

\section{Discussion and Future Work}
While this work rigorously validates the Evolutionary Forecasting (EF) paradigm on normal deep learning architectures, a compelling avenue for future research is to extend this inquiry to the emerging landscape of Time Series Foundation Models.
We hypothesize that the optimization pathology—specifically the gradient conflicts inherent in long-horizon supervision—may be a universal law that persists even in massive pre-training regimes.
Consequently, we anticipate that the ``To See Far, Look Close'' philosophy could inspire novel, computation-efficient pre-training strategies for foundation models. By shifting the focus from expensive long-sequence learning to high-fidelity short-horizon reasoning, EF holds the potential to serve as a generalized training protocol, driving the next leap in scalable time series modeling.

\end{document}